\documentclass{article}%
\usepackage{amsmath}
\usepackage{amsfonts}
\usepackage{amssymb}
\usepackage{graphicx}
\usepackage{algorithmic}
\usepackage{algorithm}%
\usepackage{hyperref}
\providecommand{\U}[1]{\protect\rule{.1in}{.1in}}
\begin{document}

\title{Interpretable Machine Learning with an Ensemble of Gradient Boosting Machines}
\author{Andrei V. Konstantinov and Lev V. Utkin\\Peter the Great St.Petersburg Polytechnic University\\St.Petersburg, Russia\\e-mail: andrue.konst@gmail.com, lev.utkin@gmail.com}
\date{}
\maketitle

\begin{abstract}
A method for the local and global interpretation of a black-box model on the
basis of the well-known generalized additive models is proposed. It can be
viewed as an extension or a modification of the algorithm using the neural
additive model. The method is based on using an ensemble of gradient boosting
machines (GBMs) such that each GBM is learned on a single feature and produces
a shape function of the feature. The ensemble is composed as a weighted sum of
separate GBMs resulting a weighted sum of shape functions which form the
generalized additive model. GBMs are built in parallel using randomized
decision trees of depth 1, which provide a very simple architecture. Weights
of GBMs as well as features are computed in each iteration of boosting by
using the Lasso method and then updated by means of a specific smoothing
procedure. In contrast to the neural additive model, the method provides
weights of features in the explicit form, and it is simply trained. A lot of
numerical experiments with an algorithm implementing the proposed method on
synthetic and real datasets demonstrate its efficiency and properties for
local and global interpretation.


\end{abstract}

\section{Introduction}

Machine learning models play an important role for making decision and
inferring predictions in various applications. However, a lot of machine
learning models are regarded as black boxes, and they cannot easily explain
their predictions or decisions in a way that humans could understand. This
fact contradicts with a requirement of understanding results provided by the
models, for example, a doctor has to have an explanation of a stated diagnosis
and has to understand why a machine learning model provides the diagnosis in
order to choose a preferable treatment \cite{Holzinger-etal-2019}, the doctor
cannot effectively use predictions provided by the models without their
interpretation and explanation. To cope with the problem of interpretation of
black-box models, a\ lot of interpretable models have been developed to
explain predictions of the deep classification and regression algorithms, for
example, deep neural network predictions
\cite{Belle-Papantonis-2020,Guidotti-2019,Molnar-2019,Xie-Ras-etal-2020}.

Interpretation methods can be divided into two groups: local and global
methods. Local methods try to explain a black-box model locally around a test
example whereas global methods derive interpretations on the whole dataset or
its part. In spite of importance and interest of the global interpretation
methods, most applications aim to understand decisions concerning with an
object to answer the question what features of the analyzed object are
responsible for a black-box model prediction. Therefore, we focus on both the
groups of interpretation methods, including local as well as global interpretations.

One of the most popular post-hoc approaches to interpretation is to
approximate a black-box model by the linear model. The well-known methods like
SHAP (SHapley Additive exPlanations)
\cite{Lundberg-Lee-2017,Strumbel-Kononenko-2010} and LIME (Local Interpretable
Model-Agnostic Explanation) \cite{Ribeiro-etal-2016} are based on building a
linear model around the instance to be explained. Coefficients of the linear
model are interpreted as the feature's importance. The linear regression for
solving the regression problem or logistic regression for solving the
classification problem allow us to construct the corresponding linear models.

However, the simple linear model cannot adequately approximate a black-box
model in some cases. Therefore, its generalization in the form of Generalized
Additive Models (GAMs) \cite{Hastie-Tibshirani-1990} is used. GAMs are a class
of linear models where the outcome is a linear combination of some functions
of features. They aim to provide a better flexibility for the approximation of
the black-box model and to determine the feature importance by analyzing how
the feature affects the predicted output through its corresponding function
\cite{Lou-etal-12,Zhang-Tan-Koch-etal-19}. GAMs can be written as follows:
\begin{equation}
y(\mathbf{x})=w_{0}+w_{1}g_{1}(x_{1})+...+w_{m}g_{m}(x_{m}).
\label{Interpr_GBM_1}%
\end{equation}
where $\mathbf{x}=(x_{1},...,x_{m})$ is the feature vector; $y$ is the target
variable; $g_{i}$ is a univariate shape function with $\mathbb{E}(g_{i})=0$;
$\mathbf{w}=(w_{1},...,w_{m})$ is the vector of coefficients.

The feature contribution to the black-box model output can be understood by
looking at the shape functions $g_{i}$ \cite{Nori-etal-19,Chang-Tan-etal-2020}.

One of the interpretation methods using GAMs is Explainable Boosting Machine
(EBM) proposed by Nori et al. \cite{Nori-etal-19}. According to EBM, shape
functions are gradient-boosted ensembles of bagged trees, each tree operating
on a single variable. Another interesting class of models called Neural
Additive Models (NAMs) was proposed by Agarwal et al. \cite{Agarwal-etal-20}.
NAMs learn a linear combination of neural networks such that a single feature
is fed to each network which performs the function $g_{i}(x_{i})$. The
networks are trained jointly. The impact of every feature on the prediction is
determined by its corresponding shape function $g_{i}$.

Similar approaches using neural networks for constructing GAMs and performing
shape functions called GAMI-Net and the Adaptive Explainable Neural Networks
(AxNNs) were proposed by Yang et al. \cite{Yang-Zhang-Sudjianto-20} and Chen
et al. \cite{Chen-Vaughan-etal-20}, respectively.

One of the popular models to explain the black-box model by means of GAMs is
the well-known gradient boosting machine (GBM)
\cite{Friedman-2001,Friedman-2002}. GBMs have illustrated their efficiency for
solving regression problems. An idea behind their using as interpretation
models is that all features are sequentially considered in each iteration of
boosting to learn shape function for all features. For example, it is shown by
Lou et al. \cite{Lou-etal-12} that all shape functions are set to zero to
initialize the algorithm. For each feature, residuals are calculated, the
one-dimensional function is learned on residuals, and it is added to the shape
function. This procedure is performed over a predefined number of iterations.
It should be noted that regression trees are often used to predict new
residuals. They are built in the GBM such that each successive tree predicts
the residuals of the preceding trees given an arbitrary differentiable loss
function \cite{Sagi-Rokach-2018}.

Another modification of the GBM for interpretation is based on the boosting
package XGBoost \cite{Chen-Guestrin-2016}. Chang et al.
\cite{Chang-Tan-etal-2020} propose to limit the tree depth to 1 to avoid
interactions of features and to convert XGBoost to the GAM.

In order to extend the set of interpretation methods using GAMs, we propose
another method based on applying GBMs with partially randomized decision trees
\cite{Konstantinov-Utkin-20a}. In contrast to extremely randomized trees
proposed by Geurts et al. \cite{Geurts-etal-06}, the cut-point for
partitioning each feature in these trees is determined randomly from the
uniform distribution, but the best feature is selected such that it maximizes
the score. The proposed method is based on a weighted sum of separate GBMs
such that each GBM depends on a single feature.

In contrast to the NAM \cite{Agarwal-etal-20}, the proposed model calculates
coefficients $w_{1},...,w_{m}$ of the GAM in the explicit form because GBMs
are built in parallel, but not sequentially. Moreover, the problem of a long
training of neural networks is partially solved because the proposed algorithm
implementing the model optimally fits coefficients $w_{1},...,w_{m}$ in each
iteration of the gradient boosting due to using the Lasso method and a
cross-validation procedure. When the weights stop changing, the training
process can be stopped, unlike EBM, where the stopping criterion is an
accuracy measure. In addition, if there exist correlations between features,
the proposed model detects them. In this case, the weights continue to grow
and will not reach a plateau.

In contrast to models proposed in \cite{Caruana-etal-2015}, the proposed
method builds many decision trees for every feature. Moreover, all GBMs are
trained simultaneously and dependently. First, coefficients of the GAM play
the role of an adaptive learning rate \cite{Konstantinov-Utkin-20a}, which
allows us to reduce the overfitting problem. Second, if the prediction is a
non-linear function, for example, in the case of classification, then
derivatives of the loss function for each GBM depend on predictions of other
GBMs. It is important to note that the GBMs in the proposed method can be used
jointly with neural networks when a part of features is processed by trees,
another part is used by networks.

A lot of numerical experiments with an algorithm implementing the proposed
method on synthetic and real datasets demonstrate its efficiency and
properties for local and global interpretation. 

The code of the proposed algorithmn can be found in 
\url{https://github.com/andruekonst/egbm-gam}.

The paper is organized as follows. Related work is in Section 2. An
introduction to the GBM, brief introductions to LIME and the NAM are provided
in Section 3 (Background). A detailed description of the proposed
interpretation method and an algorithm implementing it are provided in Section
4. Numerical experiments with synthetic data and real data using the global
interpretation are given in Section 5. Numerical experiments with the local
interpretation are given in Section 6. Concluding remarks can be found in
Section 7.

\section{Related work}

\textbf{Local and global interpretation methods. }Due to importance of the
machine learning model interpretation in many applications, a lot of methods
have been proposed to explain black-box models locally. One of the first local
interpretation methods is the Local Interpretable Model-agnostic Explanations
(LIME) \cite{Ribeiro-etal-2016}, which uses simple and easily understandable
linear models to approximate the predictions of black-box models locally.
Following the original LIME \cite{Ribeiro-etal-2016}, a lot of its
modifications have been developed due to a nice simple idea underlying the
method to construct a linear approximating model in a local area around a test
example. Examples of these modifications are ALIME
\cite{Shankaranarayana-Runje-2019}, Anchor LIME \cite{Ribeiro-etal-2018},
LIME-Aleph \cite{Rabold-etal-2019}, GraphLIME \cite{Huang-Yamada-etal-2020},
SurvLIME \cite{Kovalev-Utkin-Kasimov-20a}. Another explanation method, which
is based on the linear approximation, is the SHAP
\cite{Lundberg-Lee-2017,Strumbel-Kononenko-2010}, which takes a game-theoretic
approach for optimizing a regression loss function based on Shapley values. A
comprehensive analysis of LIME, including the study of its applicability to
different data types, for example, text and image data, is provided by Garreau
and Luxburg \cite{Garreau-Luxburg-2020}. The same analysis for tabular data is
proposed by Garreau and Luxburg \cite{Garreau-Luxburg-2020a}. An interesting
information-theoretic justification of interpretation methods on the basis of
the concept of explainable empirical risk minimization is proposed by Jung
\cite{Jung-20}.

An important group of interpretation methods is based on perturbation
techniques
\cite{Fong-Vedaldi-2019,Fong-Vedaldi-2017,Petsiuk-etal-2018,Vu-etal-2019}. The
basic idea behind the perturbation techniques is that contribution of a
feature can be determined by measuring how a prediction score changes when the
feature is altered \cite{Du-Liu-Hu-2019}. Perturbation techniques can be
applied to a black-box model without any need to access the internal structure
of the model. However, the corresponding methods are computationally complex
when samples are of the high dimensionality.

Global interpretations are more difficult tasks, and there are a few papers
devoted to solving the global interpretation tasks
\cite{Blakely-Granmo-20,Huber-etal-20,Karlsson-etal-20,Lundberg-etal-20,Mikolajczyk-etal-20,Yang-Rangarajan-Ranka-18}%
. We also have to point out models which can be used as local and global
interpretations
\cite{Agarwal-etal-20,Caruana-etal-2015,Chang-Tan-etal-2020,Chen-Vaughan-etal-20,Lou-etal-12,Nori-etal-19,Yang-Zhang-Sudjianto-20}%
.

A lot of interpretation methods, their analysis, and critical review can be
found in survey papers
\cite{Adadi-Berrada-2018,Arrieta-etal-2019,Belle-Papantonis-2020,Carvalho-etal-2019,Das-Rad-20,Guidotti-2019,Rudin-2019,Xie-Ras-etal-2020}%
.

\textbf{Gradient boosting machines with randomized decision trees}. GBMs are
efficient tool for solving regression and classification problems. Moreover,
it is pointed out by Lundberg et al. \cite{Lundberg-etal-20} that GBMs with
decision trees as basic models can be more accurate than neural networks and
more interpretable than linear models. Taking into account this advantage of
decision trees in GBMs, a modification of the GBM on the basis of the deep
forests \cite{Zhou-Feng-2017a} called the multi-layered gradient boosting
decision tree model is proposed by Feng et al. \cite{Feng-Yu-Zhou-2018}.
Another modification is the soft GBM \cite{Feng-etal-20}. It turns out that
randomized decision trees \cite{Geurts-etal-06} may significantly improve GBMs
with decision trees. As a results, several models are implemented by using
different modifications of randomized decision trees
\cite{Konstantinov-Utkin-20}. These methods can be successfully applied to the
local and global interpretation problems.

\section{Background}

\subsection{A brief introduction to the GBM for regression}

The standard regression problem can be stated as follows. Given $N$ training
examples $D=\{(\mathbf{x}_{1},y_{1}),...,(\mathbf{x}_{N},y_{N})\}$, in which
$\mathbf{x}_{i}$ belongs to a set $\mathcal{X}\subset\mathbb{R}^{m}$ and
represents a feature vector involving $m$ features, and $y_{i}\in\mathbb{R}$
represents the observed output or the target value such that $y_{i}%
=q(\mathbf{x}_{i})+\varepsilon$. Here $\varepsilon$ is the random noise with
expectation $0$ and unknown finite variance. Machine learning aims to
construct a regression model or an approximation $g$ of the function $f$ that
minimizes the expected risk or the expected loss function
\begin{equation}
L(g)=\mathbb{E}_{(\mathbf{x},y)\sim P}~l(y,g(\mathbf{x}))=\int_{\mathcal{X}%
\times\mathbb{R}}L(y,g(\mathbf{x}))\mathrm{d}P(\mathbf{x},y),
\label{Imp_SVM16}%
\end{equation}
with respect to the function parameters. Here $P(\mathbf{x},y)$ is a joint
probability distribution of $\mathbf{x}$ and $y$; the loss function
$L(\cdot,\cdot)$ may be represented, for example, as follows:
\begin{equation}
L(y,g(\mathbf{x}))=\left(  y-g(\mathbf{x})\right)  ^{2}.
\end{equation}

There are many powerful machine learning methods for solving the regression
problem, including regression random forests
\cite{Biau-Scornet-2016,Breiman-2001}, the support vector regression
\cite{Smola-Scholkopf-2004}, etc. One of the powerful methods is the GBM
\cite{Friedman-2002}, which will be briefly considered below.

GBMs iteratively improve the predictions of $y$ from $\mathbf{x}$ with respect
to $L$ by adding new weak or base learners that improve upon the previous
ones, forming an additive ensemble model of size $M$:
\begin{equation}
g_{0}(\mathbf{x})=c,\ \ g_{i}(\mathbf{x})=g_{i-1}(\mathbf{x})+\gamma_{i}%
h_{i}(\mathbf{x}),\ i=1,...,M. \label{grad_bost_10}%
\end{equation}
where $i$ is the iteration index; $h_{i}$ is the $i$-th base model, for
example, a decision tree; $\gamma_{i}$ is the coefficient or the weight of the
$i$-th base model.

The algorithm aims to minimize the loss function $L$, for example, the
squared-error $L_{2}$-loss, by iteratively computing the gradient in
accordance with the standard gradient descent method. If to say about decision
trees as the base models, then a single decision tree is constructed in each
iteration to fit the negative gradients. The function $h_{i}$ can be defined
by parameters $\theta_{i}$, i.e., $h_{i}(\mathbf{x})=h(\mathbf{x},\theta_{i}%
)$. However, we will not use the parametric representation of the function.

The gradient boosting algorithm can be represented be means of Algorithm
\ref{alg:Interpr_GBM_0}.

\begin{algorithm}
\caption{The GBM algorithm}\label{alg:Interpr_GBM_0}

\begin{algorithmic}
[1]\REQUIRE Training set $D$; the number of the GBM iterations $T$

\ENSURE Predicted function $g(\mathbf{x})$

\STATE Initialize the zero base model $g_{0}(x)$, for example, with the
constant value.

\FOR{$t=1$, $t\leq M$ }

\STATE Calculate the residual $q_{i}^{(t)}$ as the partial derivative of the
expected loss function $L(y_{i},g_{t}(\mathbf{x}_{i}))$ at every point of the
training set, $t=1,...,N$, (the negative gradient)
\begin{equation}
q_{i}^{(t)}=-\left.  \frac{\partial L(y_{i},z)}{\partial z}\right\vert
_{z=g_{i-1}(\mathbf{x}_{i})} \label{Interpr_GBM_10}%
\end{equation}

\STATE Build a new base model (a regression tree) $h_{t}(\mathbf{x}_{i})$ on
dataset $\{(\mathbf{x}_{i},q_{i}^{(t)})\}$

\STATE Find the best gradient descent step-size $\gamma_{t}$:%
\begin{equation}
\gamma_{t}=\arg\min_{\gamma}\sum_{i=1}^{N}L(y_{t},g_{t-1}(\mathbf{x}%
_{i})+\gamma h_{t}(\mathbf{x}_{i}))
\end{equation}

\STATE Update the whole model $g_{t}(\mathbf{x})=g_{t-1}(\mathbf{x}%
)+\gamma_{t}h_{t}(\mathbf{x})$;

\ENDFOR

\STATE Output
\begin{equation}
g_{M}(\mathbf{x})=\sum_{t=1}^{M}\gamma_{t}h_{t}(\mathbf{x})=g_{M-1}%
(\mathbf{x})+\gamma_{M}h_{M}(\mathbf{x}).
\end{equation}

\end{algorithmic}
\end{algorithm}

The above algorithm minimizes the expected loss function by using decision
trees as base models. Its parameters include depths of trees, the learning
rate, the number of iterations. They are selected to provide a high
generalization and accuracy depending on an specific task.

The gradient boosting algorithm is a powerful and efficient tool for solving
regression problems, which can cope with complex non-linear function
dependencies \cite{Natekin-Knoll-13}.

\subsection{Neural Additive Models}

\begin{figure}
[ptb]
\begin{center}
\includegraphics[
%%=7.820500in,
%%=9.280300in,
height=2.2407in,
width=2.6567in
]%
{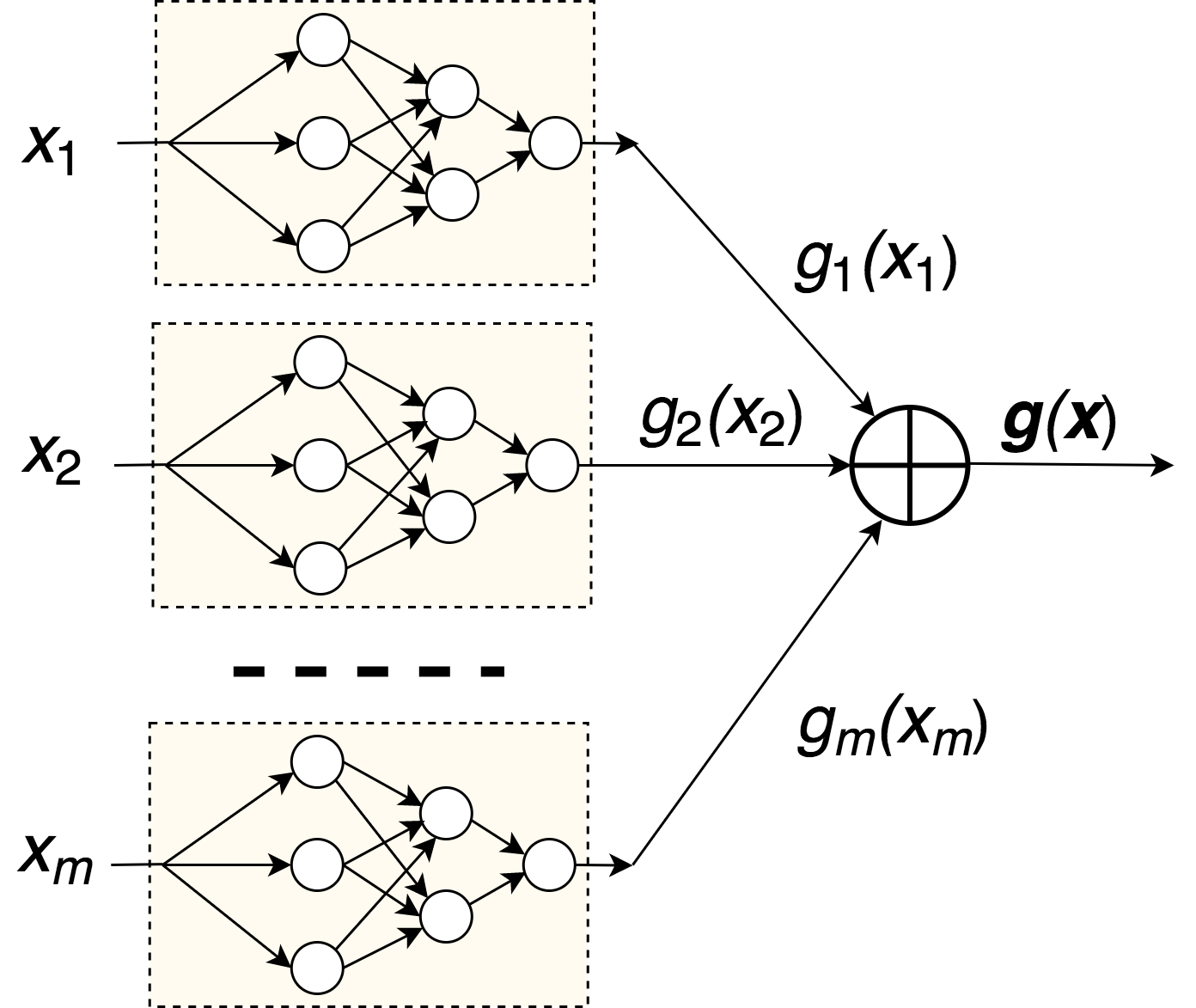}%
\caption{A scheme of the NAM given in \cite{Agarwal-etal-20}}%
\label{f:neural_n_1}%
\end{center}
\end{figure}

The proposed interpretation method can be regarded as a modification of the
NAM because this method takes some ideas behind the NAM, namely, the separate
training on single features and summing the shape functions $g_{i}$.
Therefore, we consider it in detail. According to \cite{Agarwal-etal-20}, a
general scheme of the NAM is shown in Fig. \ref{f:neural_n_1}. The separate
networks with inputs $\mathbf{x}_{1},...,\mathbf{x}_{m}$ are trained jointly
using backpropagation. They can realize arbitrary shape functions because
there are no restrictions on these networks. However, a difficulty of learning
the neural network may be a possible small amount of training data in case of
the global interpretation. In this case, the network may be overfitted. On the
one hand, to cope with the overfitting problem, the separate networks can be
constructed as small as possible to reduce the number of their training
parameters. On the other hand, simple neural networks may have difficulties in
realizing the corresponding shape functions.%

%TCIMACRO{\FRAME{ftbpFU}{2.6567in}{2.2407in}{0pt}{\Qcb{A scheme of the NAM
%given in \cite{Agarwal-etal-20}}}{\Qlb{f:neural_n_1}}{neural_n_1.png}%
%width 2.6567in;  height 2.2407in;  depth 0pt;  original-width 9.2803in;
%original-height 7.8205in;  cropleft "0";  croptop "1";  cropright "1";
%cropbottom "0";  filename 'Neural_N_1.png';file-properties "XNPEU";}}
%}%
%BeginExpansion
%EndExpansion

\subsection{LIME}

It is important to point out that methods based on using GAMs and NAMs
\cite{Agarwal-etal-20,Caruana-etal-2015,Chang-Tan-etal-2020,Chen-Vaughan-etal-20,Lou-etal-12,Nori-etal-19,Yang-Zhang-Sudjianto-20}
as well as the proposed method can be regarded as some modifications of LIME
for local interpretation to some extent. In the case of local interpretation,
the methods have to use the perturbation technique and to minimize the loss
function which measures how the interpretation is close to the prediction. The
same elements are used in LIME \cite{Ribeiro-etal-2016}. Therefore, we briefly
consider the original LIME.

LIME aims to approximate a black-box interpretable model denoted as $f$ with a
simple linear function $g$ in the vicinity of the point of interest
$\mathbf{x}$, whose prediction by means of $f$ has to be interpreted, under
condition that the approximation function $g$ belongs to a set of linear
functions $G$. According to LIME, a new dataset consisting of perturbed
examples $\mathbf{x}_{k}$ around point $\mathbf{x}$ is generated to construct
the function $g$, and predictions corresponding to the perturbed examples are
obtained by means of the black-box model, i.e., by computing $f(\mathbf{x}%
_{k})$ for many points $\mathbf{x}_{k}$. Weights $w_{\mathbf{x}}$ are assigned
to new examples in accordance with their proximity to the point of interest
$\mathbf{x}$ by using a distance metric, for example, the Euclidean distance.

An interpretation (local surrogate) model is trained on new generated samples
by solving the following optimization problem:
\begin{equation}
\arg\min_{g\in G}L(f,g,w_{\mathbf{x}})+\Phi(g).
\end{equation}

Here $L$ is a loss function, for example, mean squared error, which measures
how the model $g$ is close to the prediction of the black-box model $f$;
$\Phi(g)$ is the regularization term.

The obtained local linear model $g$ interprets the prediction by analyzing
coefficients of $g$. It should be noted that LIME can also apply GAMs as an
extension of $G$. However, the shape functions of features have to be known
for constructing the optimization problem.

\section{The interpretation method and the algorithm for regression}

The basic idea behind the proposed interpretation method is to realize a
weighted sum of GBMs such that each GBM depends on a single feature. Moreover,
weights of features obtained by means of the proposed algorithm can be used
for interpretation because their absolute values can be viewed as impacts of
the features.

The proposed method of interpretation can play roles of the \textit{local}
interpretation when a test example $\mathbf{x}$ is interpreted and a new
dataset consisting of perturbed samples at a local area around the test
example $\mathbf{x}$ is generated, as well as the \textit{global}
interpretation when the interpretation model is built on the entire dataset.
Plots showing the corresponding shape functions $g_{k}$ of features describe
the model behavior. The difference between the local and global models from
the implementation point of view is in a dataset for training the proposed
interpretation model. In particular, outcomes $y_{i}$ of the black-box model
in the local case are values of $f(\mathbf{x}_{i})$ obtained for examples
$\mathbf{x}_{i}$ generated in a local area around the test example
$\mathbf{x}$. In case of the global interpretation, outcomes $y_{i}$ of the
black-box model are values of $f(\mathbf{x}_{i})$ obtained for examples from
the initial dataset.

The algorithm implementing the proposed interpretation method for regression
is based on the iterative use of a GBM ensemble consisting of $m$ parallel
GBMs such that each GBM deals with a separate feature. The corresponding model
is composed as a weighted sum of separate GBMs such that each GBM depends on a
single feature. Partially randomized decision trees
\cite{Konstantinov-Utkin-20a} are used for implementing each GBM. The
cut-point for partitioning each feature in these trees is randomly selected
from the uniform distribution, but the optimal feature is selected to maximize
the score. The tree depth is limited to 1. It is important to point out that
arbitrary trees can be used for implementing the GBMs. However, various
experiments show that the used depth 1 significantly reduces the risk of
overfitting. Moreover, the training time is significantly reduced when the
depth of trees is 1. It is a very interesting and unexpected observation that
trees of depth 1 with the random splitting of each feature provide better
results than deep trees. This observation again illustrates the strength of
the GBM as a machine learning model.

Initially, each GBM computes functions $g^{(s)}(x_{i,k})$ of the $k$-th
feature in the $s$-th iteration of the whole algorithm by using residuals
$r_{i}^{(s)}\cdot w_{k}^{(s)}$, $i=1,...,N$, obtained from all training
examples, and the feature vectors $\mathbf{x}_{i}$, $i=1,...,N$, as input
data. Here $w_{k}^{(s)}$ is the weight of the $k$-th feature. The upper index
$s$ shows the iteration number of the ensemble. It is assumed that all weights
are initially equal to 1, i.e., $w_{k}^{(0)}=1$ for $k=1,...,m$. In other
words, the input data is the set of pairs $(x_{i,k},r_{i}^{(s)})$,
$i=1,...,N$. If to initialize residuals as $r_{i}^{(s)}=0$, then the iteration
with $s=0$ starts with the training set $(x_{i,k},0)$, $i=1,...,N$. One should
distinguish residuals $q_{i}^{(t)}$ obtained in each GBM (\ref{Interpr_GBM_10}%
) (see Algorithm \ref{alg:Interpr_GBM_0}) and residuals $r_{i}^{(s)}$ obtained
in the ensemble of GBMs. Moreover, in order to avoid misunderstanding, we will
consider every GBM based on decision trees and predicting a separate feature
as a minimal element without its detal.

\begin{figure}
[ptb]
\begin{center}
\includegraphics[
%%=7.260100in,
%%=21.239799in,
height=2.1871in,
width=6.3693in
]%
{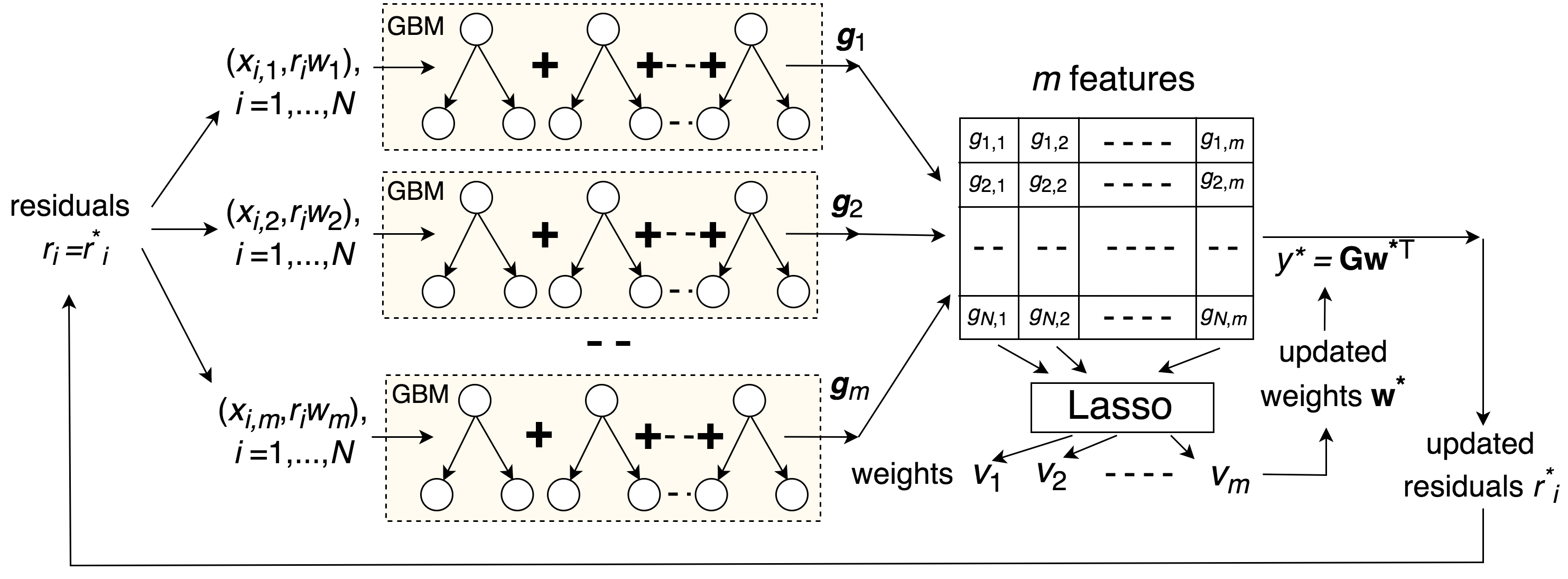}%
\caption{A scheme of the proposed interpretation algorithm (the EGBM)}%
\label{f:Interpr_GBM_2}%
\end{center}
\end{figure}

A general scheme of the algorithm implementing the proposed interpretation
method is shown in Fig. \ref{f:Interpr_GBM_2}. Let us denote the
$N$-dimensional vector of functions $g^{(s)}(x_{i,k})$ for all examples from
the dataset computed by the $k$-th GBM as $\mathbf{g}_{k}^{(s)}$, and the
$N\times m$ matrix of functions $g^{(s)}(x_{i,k})$ as $\mathbf{G}^{(s)}$,
where $\mathbf{G}^{(s)}=[\mathbf{g}_{1}^{(s)},...,\mathbf{g}_{m}^{(s)}]$. The
importance of features can be estimated by computing weights $w_{k}^{(s)}$,
$k=1,...,m$, of every shape function or every vector $\mathbf{g}_{k}^{(s)}$.
In order to assign weights to vectors $\mathbf{g}_{k}^{(s)}$, $k=1,...,m$, the
Lasso method \cite{Tibshirani-1996} is used. It should be noted that other
methods for assigning the weights can be used, for example, the ridge
regression or the elastic net regression. However, our numerical experiments
show that the Lasso method significantly reduces time (and the number of the
ensemble iterations) for convergence of the weights. Denote weights computed
by using the Lasso method as $v_{k}^{(s)}$. In order to smooth the weight
changes in each iteration, we propose to update the weights by means of the
following rule:
\begin{equation}
w_{k}^{(s)}=(1-\alpha)w_{k}^{(s-1)}+\alpha\cdot v_{k}^{(s)},\ k=1,...,m.
\label{Interpr_GBM_34}%
\end{equation}

Here $\alpha\in(0,1]$ is the smoothing parameter. In particular, if $\alpha
=1$, then \textquotedblleft old\textquotedblright\ weights $w_{k}^{(s-1)}$
obtained in the previous iteration are not used and weights $w_{k}^{(s)}$ are
totally determined by the current iteration. This case may lead to a better
convergence of weights due to their possible large deviations.

Having the updated weights, new target values $y_{i}^{\ast(s)}$, $i=1,...,N$,
are computed by multiplying matrix $\mathbf{G}^{(s)}$ by vector $[w_{1}%
^{(s)},...,w_{m}^{(s)}]^{\mathrm{T}}$. The values $y_{i}^{\ast(s)}$ can be
viewed as current predictions of the model after $s$ iterations. Hence, the
residual $r_{i}^{(s)}$ can be computed as the partial derivative (the negative
gradient) of the expected loss function $L(y_{i},y_{i}^{\ast(s)})$ at every
point of the training set, $t=1,...,N$,
\begin{equation}
r_{i}^{(s)}=-\left.  \frac{\partial L(y_{i},z)}{\partial z}\right\vert
_{z=y_{i}^{\ast(s)}}. \label{Interpr_GBM_36}%
\end{equation}

In contrast to the standard GBM, we have to get residual $r_{i,k}^{(s)}$ for
every example as well as every feature because every feature is separately
processed by the corresponding GBM. It should be noted that there holds%
\begin{equation}
y_{i}^{\ast(s)}=\sum_{k=1}^{m}g^{(s)}(x_{i,k})w_{k}^{(s)}.
\end{equation}
Hence, we get
\begin{align}
r_{i,k}^{(s)}  &  =-\left.  \frac{\partial L\left(  y_{i},\sum_{k=1}^{m}%
z_{ik}w_{k}^{(s)}\right)  }{\partial z_{ik}}\right\vert _{z_{ik}%
=g^{(s)}(x_{i,k})}\nonumber\\
&  =-\left.  \frac{\partial L(y_{i},z)}{\partial z}\right\vert _{z=y_{i}%
^{\ast(s)}}\cdot\left.  \frac{\partial z}{\partial t}\right\vert
_{t=g^{(s)}(x_{i,k})}\nonumber\\
&  =r_{i}^{(s)}\cdot w_{k}^{(s)}. \label{Interpr_GBM_38}%
\end{align}

The above implies that the training set for learning the $k$-th GBM at the
next iteration is represented by pairs $(x_{i,k},r_{i}^{(s)}\cdot w_{k}%
^{(s)})$, $i=1,...,N$. The same training sets are used by all GBMs which
correspond to features with indices $k=1,...,m$.

One of the problems is that the obtained weights may be of different scales.
In order to overcome this problem and to correct the weights, coefficients are
multiplied by the standard deviation of every feature computed with using all
points from the dataset. The obtained corrected weights will be viewed as the
feature importance. The same procedure has been performed by Chen et al.
\cite{Chen-Vaughan-etal-20}.

Several rules for stopping the algorithm can be proposed. The algorithm can
stop iterations when the weights do not change or they insignificantly change
with some relative deviation $\epsilon$. However, this rule may lead to an
enormous number of iteration when weights are not stabilized due to
overfitting problems. Therefore, we use just some predefined value $T$ of
iterations, which can be tuned in accordance with the weight changes.

Algorithm \ref{alg:Interpr_GBM_1} can be viewed as a formal scheme for
computing the weights and shape functions.

\begin{algorithm}
\caption{The interpretation algorithm for regression}\label{alg:Interpr_GBM_1}

\begin{algorithmic}
[1]\REQUIRE Training set $D$; point of interest $\mathbf{x}$; the number of
generated or training points $N$; the smoothing parameter $\alpha$; the
black-box survival model for explaining $f(\mathbf{x})$; the number of
iterations $T$

\ENSURE Vector of weights $w_{k}$; shape functions $g_{k}(x)$ for every feature

\STATE Initialize weights $w_{k}^{(0)}=1$, $k=1,...,m$

\STATE Initialize residuals $r_{i,k}^{(0)}=0$, $k=1,...,m$, $i=1,...,N$

\STATE Standardize $y_{i}$, $i=1,...,N$

\STATE Generate $N-1$ random nearest points $\mathbf{x}_{k}$ in a local area
around $\mathbf{x}$, point $\mathbf{x}$ is the $N$-th point

\FOR{$s=0$, $s\leq T$ }

\FOR{$k=1$, $k\leq m$ }

\STATE Learn the GBM with partially randomized decision trees of depth $1$ on
the dataset $\left(  x_{i,k},r_{i,k}^{(s)}w_{k}^{(s)}\right)  $, $i=1,...,N$

\STATE Predict vector $\mathbf{g}_{k}^{(s)}$ of shape functions

\ENDFOR

\STATE Compute weights $v_{k}^{(s)}$ by using the Lasso method trained on
matrix $\mathbf{G}^{(s)}=[\mathbf{g}_{1}^{(s)},...,\mathbf{g}_{m}^{(s)}]$ as
the training feature vectors and on residuals $r_{1}^{(s-1)},...,r_{N}%
^{(s-1)}$ as predictions

\STATE Update weights $w_{k}^{(s)}$, $k=1,...,m$, by means of
(\ref{Interpr_GBM_34})

\STATE Compute $r_{i}^{(s)}$ and $r_{i,k}^{(s)}$ by using
(\ref{Interpr_GBM_36}) and (\ref{Interpr_GBM_38}), respectively, $k=1,...,m$,
$i=1,...,N$

\ENDFOR

\STATE Correct weights $w_{k}^{(T)}$, $k=1,...,m$, by using the standard deviation

\STATE$w_{k}=w_{k}^{(T)}$; $g_{k}(x)=\mathbf{g}_{k}^{(T)}$, $k=1,...,m$
\end{algorithmic}
\end{algorithm}

\begin{figure}
[ptb]
\begin{center}
\includegraphics[
%%=4.039500in,
%%=18.679899in,
height=1.3292in,
width=6.0891in
]%
{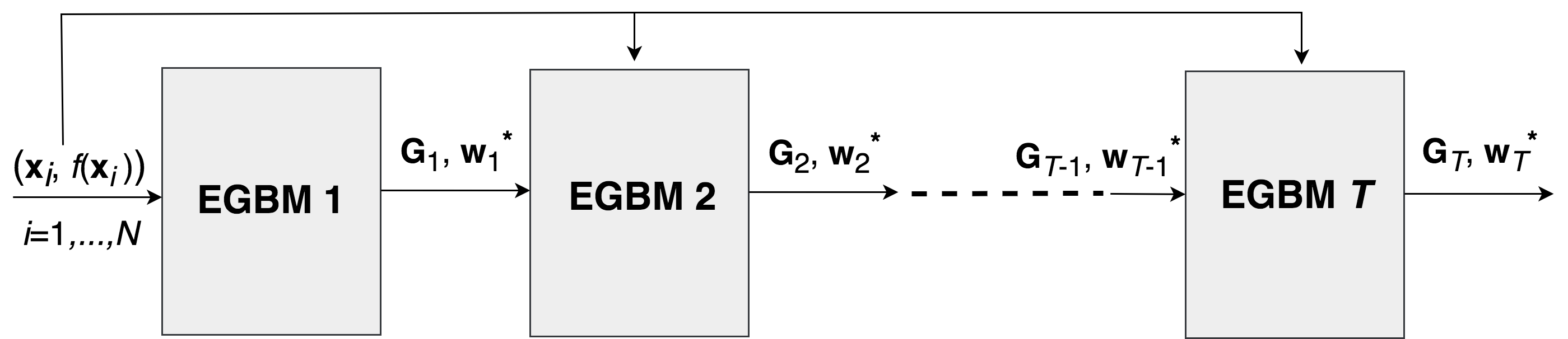}%
\caption{An outline scheme of the \textquotedblleft global\textquotedblright%
\ boosting algorithm}%
\label{f:Interpr_GBM_3}%
\end{center}
\end{figure}

In fact, we have the gradient boosting algorithms in the \textquotedblleft
global\textquotedblright\ boosting algorithm which computes weights and shape
functions $w_{k}=w_{k}^{(T)}$; $g_{k}(x)=g_{k}^{(T)}$ for every feature. An
outline scheme of this \textquotedblleft global\textquotedblright\ boosting
algorithm is shown in Fig. \ref{f:Interpr_GBM_3}. It conditionally can be
represented as a series of the gradient boosting machine ensembles (EGBMs),
where every EGBM is the algorithm given in Fig. \ref{f:Interpr_GBM_2}. Each
EGBM uses the initial dataset for training in the $s$-th iteration as well as
matrix $\mathbf{G}^{(s-1)}$ of the shape function values and weights
$w_{1}^{(s-1)},...,w_{m}^{(s-1)}$ computed in the previous iteration.%

%TCIMACRO{\FRAME{ftbpFU}{6.3693in}{2.1871in}{0pt}{\Qcb{A scheme of the proposed
%interpretation algorithm (the EGBM)}}{\Qlb{f:Interpr_GBM_2}}{gbm_1.png}%
%width 6.3693in;  height 2.1871in;  depth 0pt;  original-width 21.2398in;
%original-height 7.2601in;  cropleft "0";  croptop "1";  cropright "1";
%cropbottom "0";  filename 'GBM_1.png';file-properties "XNPEU";}} }%
%BeginExpansion
%EndExpansion
%

%TCIMACRO{\FRAME{ftbpFU}{6.0891in}{1.3292in}{0pt}{\Qcb{An outline scheme of the
%\textquotedblleft global\textquotedblright\ boosting algorithm}}%
%{\Qlb{f:Interpr_GBM_3}}{gbm_2.png}{\special{ language "Scientific Word";
%type "GRAPHIC";  maintain-aspect-ratio TRUE;  display "USEDEF";
%valid_file "F";  width 6.0891in;  height 1.3292in;  depth 0pt;
%original-width 18.6799in;  original-height 4.0395in;  cropleft "0";
%filename 'GBM_2.png';file-properties "XNPEU";}} }%
%BeginExpansion
%EndExpansion

The number of iterations $T$ is chosen large enough. It will be illustrated
below by means of numerical examples that a lack of the weight $w_{k}^{(s)}$
stabilization means that there is a strong correlation between features, which
leads to overfitting of the GBMs.

The interpretation algorithm for classification is similar to the studied
above algorithm. However, in contrast to the regression problem, we have
softmax function and the cross-entropy loss for computing residuals. Moreover,
it also makes sense to pre-train each GBM in classification to predict the
target value by means of performing several iterations. The pre-training
iterations aim to separate points by a hyperplane at the beginning of the
training process, otherwise all points may be equal to $0$.

\section{Numerical experiments using the local interpretation}

\subsection{Numerical experiments with synthetic data}

In order to study the proposed interpretation method, we consider several
numerical examples for which training examples are randomly generated.

\subsubsection{Linear regression function}

Let us suppose that the studied regression function is known, and it is
\begin{equation}
y(\mathbf{x})=10x_{1}-20x_{2}-2x_{3}+3x_{4}+0x_{5}+0x_{6}+0x_{7}+\varepsilon.
\label{Interpr_GBM_60}%
\end{equation}

\begin{figure}
[ptb]
\begin{center}
\includegraphics[
%%=5.369600in,
%%=10.406300in,
height=1.8665in,
width=3.6079in
]%
{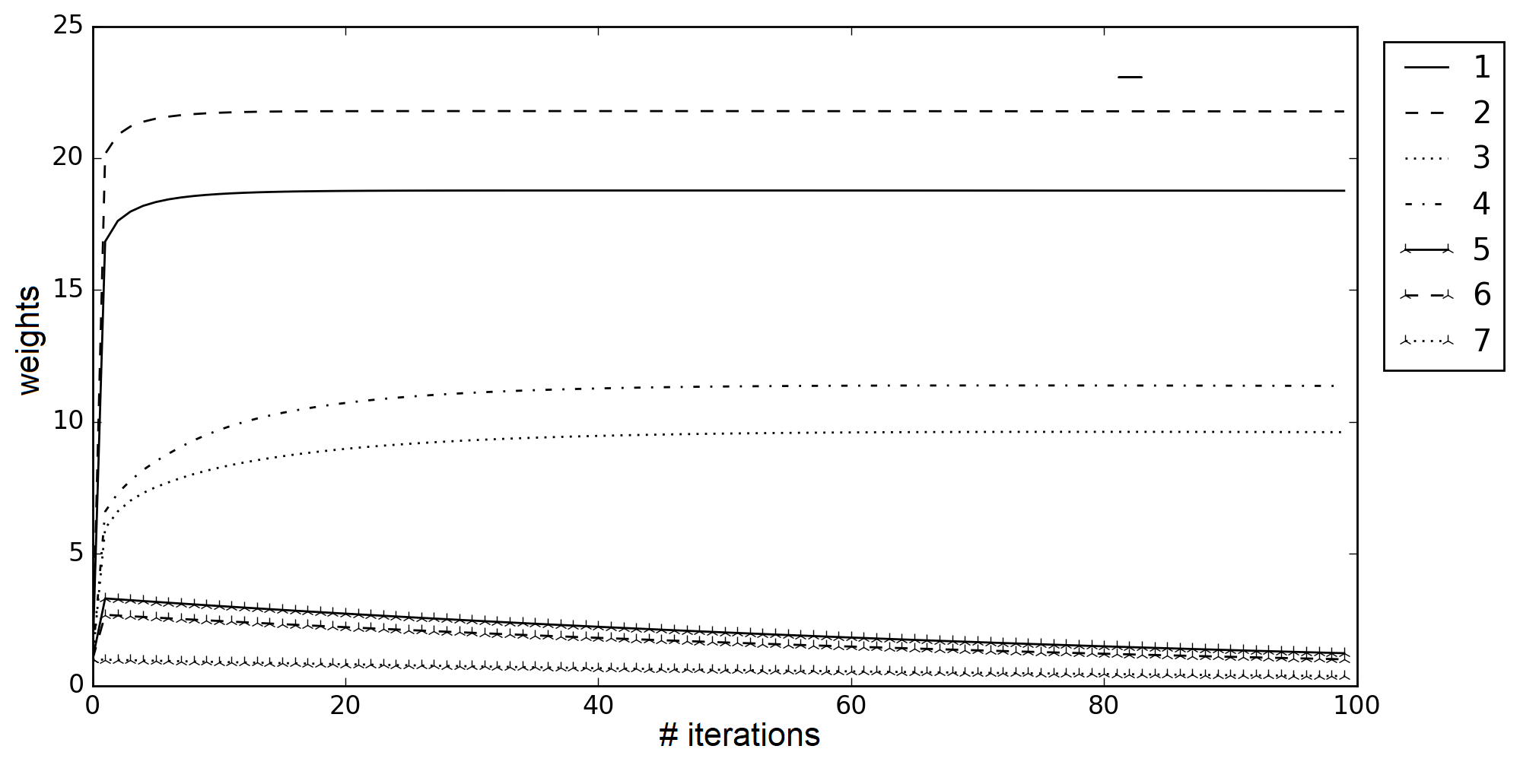}%
\caption{Weights of features as functions of the iteration number for the
linear regression}%
\label{f:history_linear}%
\end{center}
\end{figure}

\begin{figure}
[ptb]
\begin{center}
\includegraphics[
%%=6.999800in,
%%=9.999800in,
height=2.3661in,
width=3.3745in
]%
{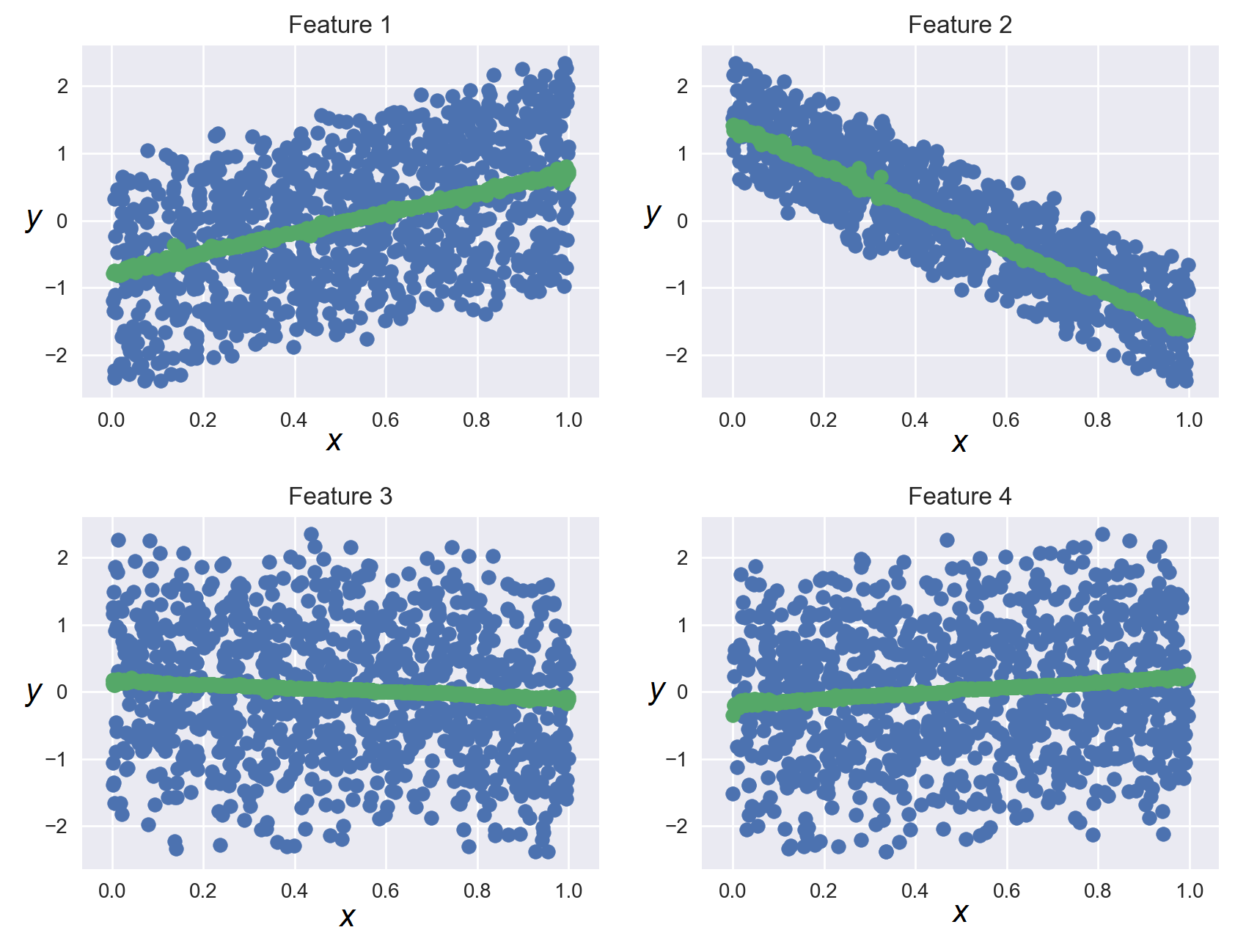}%
\caption{Predictions of single GBMs as functions of the corresponding feature
values}%
\label{f:featurewise_linear}%
\end{center}
\end{figure}

There are $7$ features whose impacts should be estimated. We generate $N=1000$
points of $\varepsilon$ with expectation $0$ and standard deviation $0.05$. It
is assumed that outputs of the black-box model correspond to these points. We
expect to get the same relationship between weights obtained by means of the
explanation model and coefficients of the above linear regression function.
Fig. \ref{f:history_linear} illustrates how the weights are changed with
increase of the number of iterations $T$. One can see that the weights tend to
converge. This is a very important property which means that there is no
overfitting of the explanation model. Moreover, one can see from Fig.
\ref{f:history_linear} that the weights totally correspond to coefficients of
the linear regression. In particular, weights of features 5, 6, 7 tend to 0,
weights of features 1, 2, 3, 4 tend to $18.77$, $21.77$, $9.16$, $11.37$,
respectively. Fig. \ref{f:featurewise_linear} shows changes of every feature,
where large points (the wide band) are generated points of the dataset. They
can be regarded as true values. The narrow band in the middle of the wide band
corresponds to predictions of a single GBM in accordance with the feature
which is fed to this GBM. The $x$-axis corresponds to the feature values, the
$y$-axis is predictions of a single GBM.

The obtained weights are $(18.77,21.77,9.16,11.37,1.24,1.01,0.37)$. These
weights are given before their correction. After the correction, we get the
feature importance values $(0.42,0.86,0.08,0.12,0,0,0)$. It can be seen from
the numerical results that relationships between interpreted weights and
between coefficients of the regression function are quite the same. This
implies that the proposed method provides correct interpretation results for
the linear function.%

%TCIMACRO{\FRAME{ftbpFU}{3.6079in}{1.8665in}{0pt}{\Qcb{Weights of features as
%functions of the iteration number for the linear regression}}%
%{\Qlb{f:history_linear}}{history_linear.png}%
%width 3.6079in;  height 1.8665in;  depth 0pt;  original-width 10.4063in;
%original-height 5.3696in;  cropleft "0";  croptop "1";  cropright "1";
%filename 'history_linear.png';file-properties "XNPEU";}} }%
%BeginExpansion
%EndExpansion
%

%TCIMACRO{\FRAME{ftbpFU}{3.3745in}{2.3661in}{0pt}{\Qcb{Predictions of single
%GBMs as functions of the corresponding feature values}}%
%{\Qlb{f:featurewise_linear}}{featurewise_linear.png}%
%width 3.3745in;  height 2.3661in;  depth 0pt;  original-width 9.9998in;
%original-height 6.9998in;  cropleft "0";  croptop "1";  cropright "1";
%filename 'featurewise_linear.png';file-properties "XNPEU";}} }%
%BeginExpansion
%EndExpansion

\subsubsection{Non-linear regression function}

Let us return to the previous numerical example, but the seventh term in
(\ref{Interpr_GBM_60}) is $g_{7}(x)=100\left(  x_{7}-0.5\right)  ^{2}$ instead
of $0x_{7}$ now. We again generate $N=1000$ points with the same parameters of
the normal distribution for $\varepsilon$ (expectation $0$ and standard
deviation $0.05$).

\begin{figure}
[ptb]
\begin{center}
\includegraphics[
%%=4.921600in,
%%=9.832900in,
height=1.9035in,
width=3.7898in
]%
{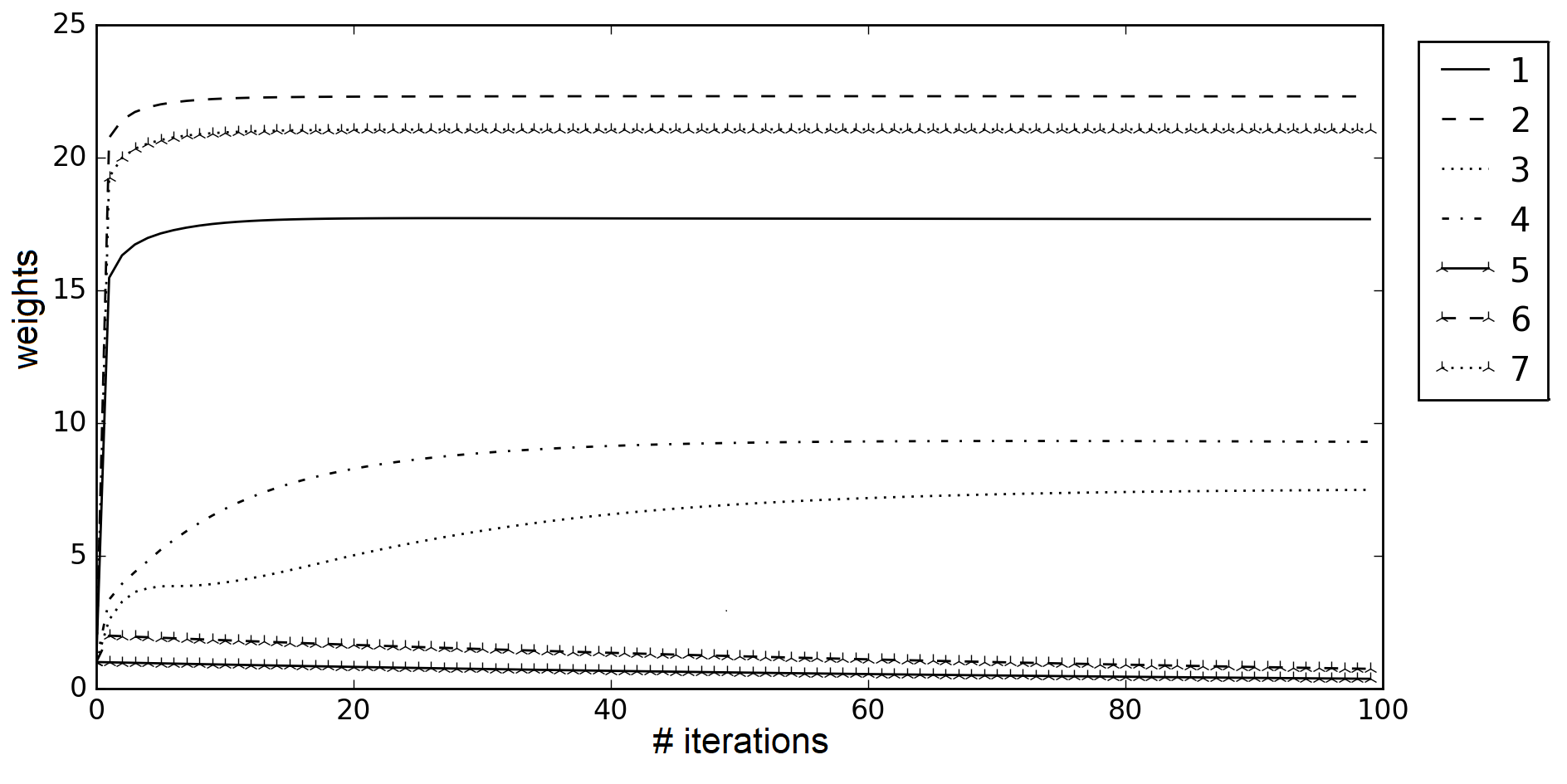}%
\caption{Weights of features as functions of the iteration number for the
non-linear regression}%
\label{f:history_nonlinear}%
\end{center}
\end{figure}

\begin{figure}
[ptb]
\begin{center}
\includegraphics[
%%=6.749100in,
%%=9.369600in,
height=2.5356in,
width=3.5137in
]%
{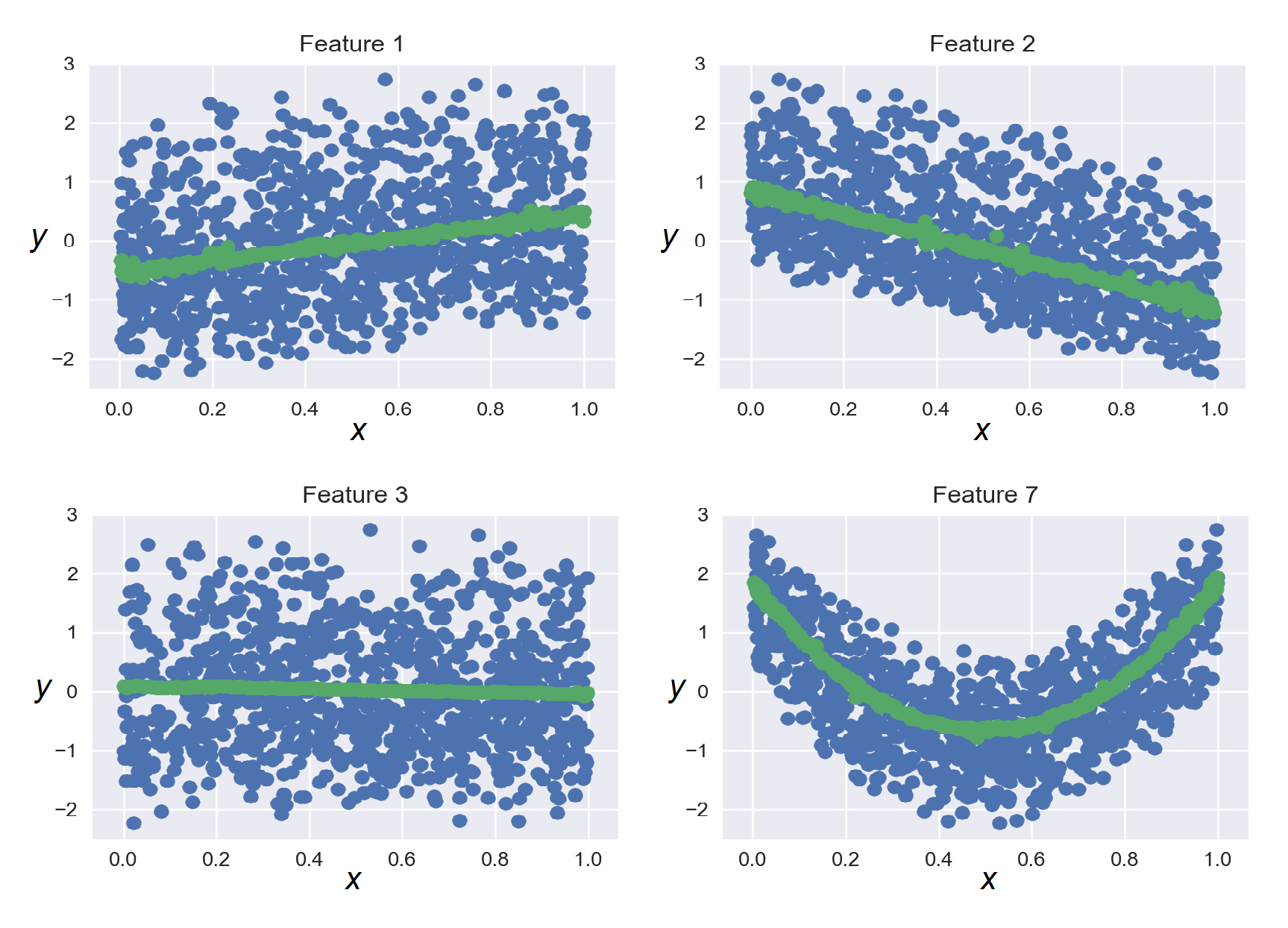}%
\caption{Predictions of single GBMs as functions of the corresponding feature
values}%
\label{f:featurewise_nonlinear}%
\end{center}
\end{figure}

The obtained weights are $(17.68,22.31,7.49,9.29,0.37,0.74,21.07)$. These
weights are presented before their correction. After the correction, we get
the feature importance values $(0.28,0.58,0.04,0.07,0,0,0.75)$. It can be seen
from the numerical results that the seventh feature $x_{7}$ (its function
$g_{7}(x)$) has the largest weight. Moreover, relationships between
interpreted weights and between coefficients of the regression function are
quite the same. This implies that the proposed method provides correct
interpretation results. It is interesting to point out that the interpretation
linear model used, for example, in LIME, could not detect the non-linearity in
many cases. Fig. \ref{f:history_nonlinear} shows how the weights are changed
with increase of the number of iterations $T$. We again observe the
convergence of weights to values given above. Fig.
\ref{f:featurewise_nonlinear} is similar to Fig. \ref{f:featurewise_linear},
and we clearly observe how the seventh feature impacts on predictions.%

%TCIMACRO{\FRAME{ftbpFU}{3.7898in}{1.9035in}{0pt}{\Qcb{Weights of features as
%functions of the iteration number for the non-linear regression}%
%}{\Qlb{f:history_nonlinear}}{history_nonlinear.png}%
%width 3.7898in;  height 1.9035in;  depth 0pt;  original-width 9.8329in;
%original-height 4.9216in;  cropleft "0";  croptop "1";  cropright "1";
%filename 'history_nonlinear.png';file-properties "XNPEU";}} }%
%BeginExpansion
%EndExpansion
%

%TCIMACRO{\FRAME{ftbpFU}{3.5137in}{2.5356in}{0pt}{\Qcb{Predictions of single
%GBMs as functions of the corresponding feature values}}%
%{\Qlb{f:featurewise_nonlinear}}{featurewise_nonlinear.png}%
%width 3.5137in;  height 2.5356in;  depth 0pt;  original-width 9.3696in;
%original-height 6.7491in;  cropleft "0";  croptop "1";  cropright "1";
%filename 'featurewise_nonlinear.png';file-properties "XNPEU";}} }%
%BeginExpansion
%EndExpansion

\subsubsection{The chess board}

\begin{figure}
[ptb]
\begin{center}
\includegraphics[
%%=4.651000in,
%%=4.843800in,
height=2.3687in,
width=2.4673in
]%
{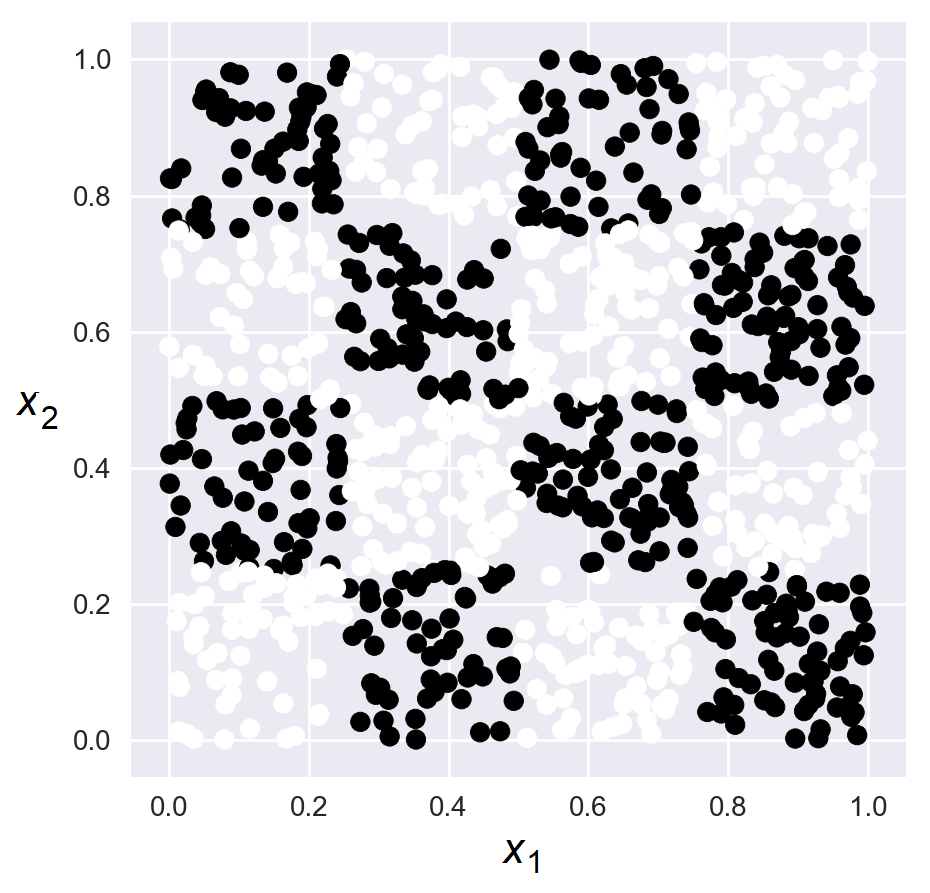}%
\caption{Generated points from two classes forming the chess board}%
\label{f:checkerboard}%
\end{center}
\end{figure}

\begin{figure}
[ptb]
\begin{center}
\includegraphics[
%%=5.159100in,
%%=9.353400in,
height=2.0444in,
width=3.6945in
]%
{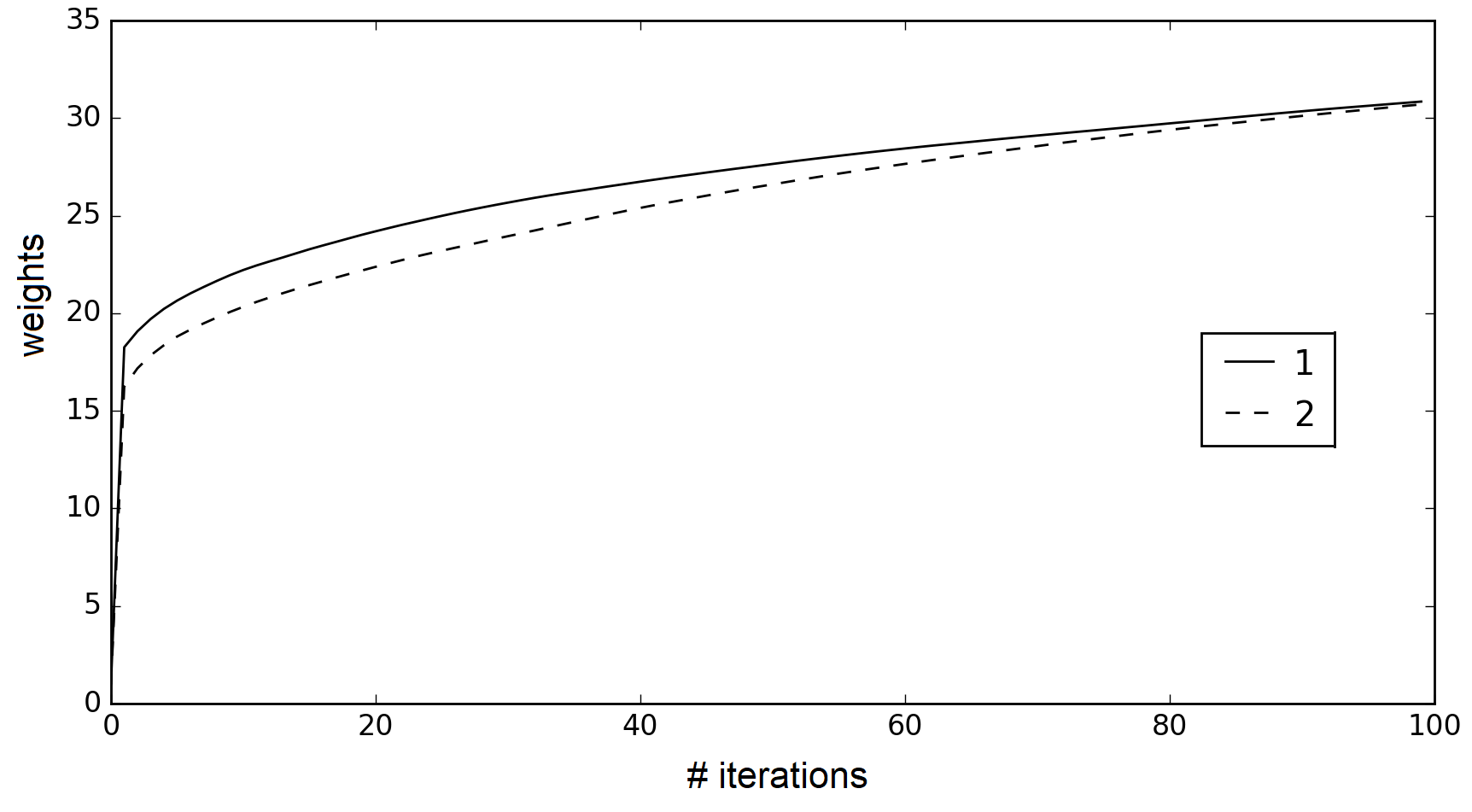}%
\caption{Weights of features as functions of the iteration number for the
chess board example}%
\label{f:history_checkerboard}%
\end{center}
\end{figure}

\begin{figure}
[ptb]
\begin{center}
\includegraphics[
%%=4.828200in,
%%=8.770900in,
height=1.6053in,
width=2.9088in
]%
{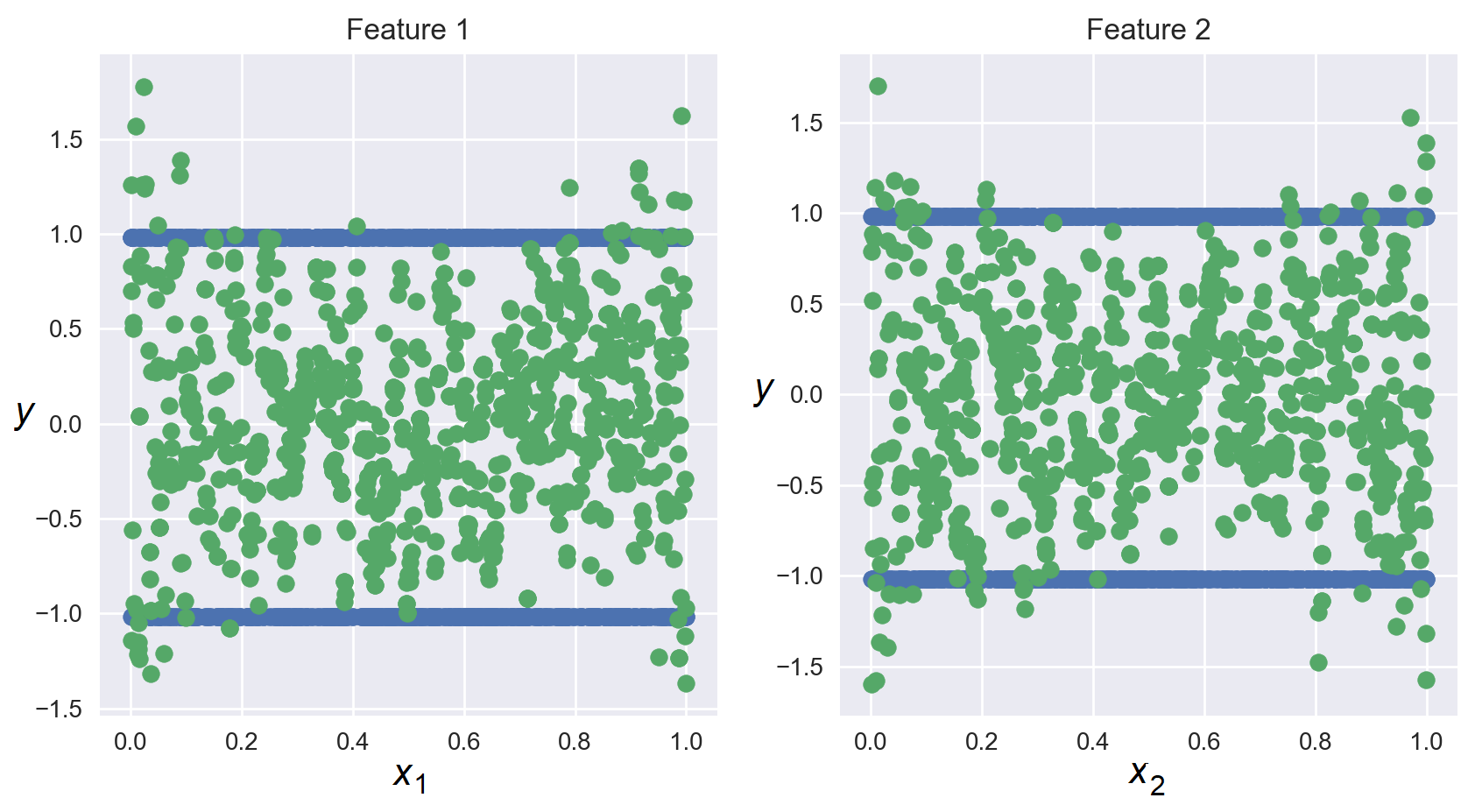}%
\caption{Predictions of GBMs as functions of the corresponding feature values}%
\label{f:featurewise_checkerboard}%
\end{center}
\end{figure}

\begin{figure}
[ptb]
\begin{center}
\includegraphics[
%%=4.588700in,
%%=4.739200in,
height=2.5062in,
width=2.5884in
]%
{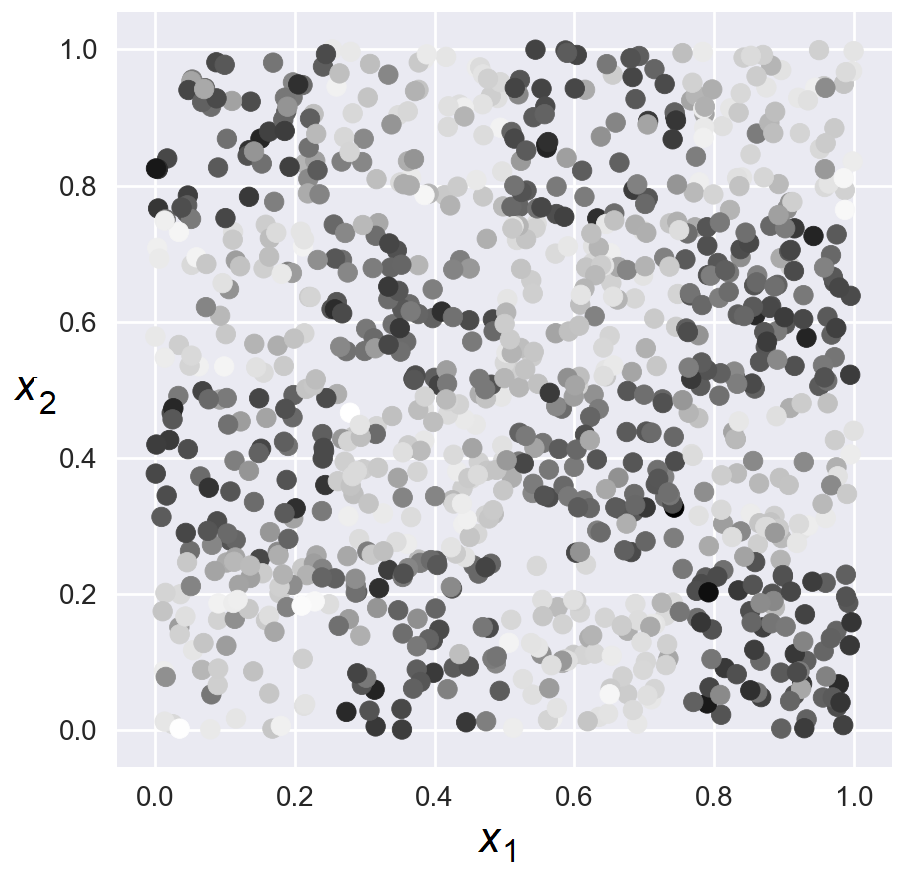}%
\caption{Predictions of the overfitted model}%
\label{f:predictions_checkerboard}%
\end{center}
\end{figure}

So far, we have considered interpretation of the regression models. The next
numerical example illustrates the binary classification task. Training
examples having two labels $-1$, $1$ and consisting of two features are
generated such that the corresponding points depict a small chess board as it
is shown in Fig. \ref{f:checkerboard}. Black and white points belong to
classes $y=1$ and $y=-1$, respectively. This is a very interesting example
which illustrates inability of the proposed algorithm to correctly classify
the given dataset due to overfitting of GBMs. Indeed, Fig.
\ref{f:history_checkerboard} shows that weights of features are not converged
and are not stabilized with increase of the iteration number $T$. The
overfitting problem arises because there is a strong correlation between
features. The same can be seen in Fig. \ref{f:featurewise_checkerboard}, where
outputs of the black-box model are depicted in the form of two lines with
$y=1$ and $y=-1$. Outcomes of the proposed algorithm are irregularly located
because the algorithm is subjected to overfitting. It is also interesting to
note that the points are scattered such that the sum of GBM outcomes for all
features provides correct values. It should also be pointed out that the
standard GBM with decision trees of depth $1$ cannot solve this problem. This
implies that the proposed parallel architecture of the tree training is better
in comparison with the serial architecture. Fig.
\ref{f:predictions_checkerboard} illustrates predicted values of the trained
explanation model.

The computed weights of features before their correction are $(29.21,30.46)$.
After correction, they are $(0.52,0.52)$. In this numerical example, weights
do not have a specific sense, but one can see that the correction method
itself provides accurate results because both the features are identical and
equal to $0.52$.%

%TCIMACRO{\FRAME{ftbpFU}{2.4673in}{2.3687in}{0pt}{\Qcb{Generated points from
%two classes forming the chess board}}{\Qlb{f:checkerboard}}{checkerboard.png}%
%width 2.4673in;  height 2.3687in;  depth 0pt;  original-width 4.8438in;
%original-height 4.651in;  cropleft "0";  croptop "1";  cropright "1";
%filename 'checkerboard.png';file-properties "XNPEU";}} }%
%BeginExpansion
%EndExpansion
%

%TCIMACRO{\FRAME{ftbpFU}{3.6945in}{2.0444in}{0pt}{\Qcb{Weights of features as
%functions of the iteration number for the chess board example}}%
%{\Qlb{f:history_checkerboard}}{history_checkerboard.png}%
%width 3.6945in;  height 2.0444in;  depth 0pt;  original-width 9.3534in;
%original-height 5.1591in;  cropleft "0";  croptop "1";  cropright "1";
%filename 'history_checkerboard.png';file-properties "XNPEU";}} }%
%BeginExpansion
%EndExpansion
%

%TCIMACRO{\FRAME{ftbpFU}{2.9088in}{1.6053in}{0pt}{\Qcb{Predictions of GBMs as
%functions of the corresponding feature values}}%
%{\Qlb{f:featurewise_checkerboard}}{featurewise_checkerboard.png}%
%width 2.9088in;  height 1.6053in;  depth 0pt;  original-width 8.7709in;
%original-height 4.8282in;  cropleft "0";  croptop "1";  cropright "1";
%filename 'featurewise_checkerboard.png';file-properties "XNPEU";}} }%
%BeginExpansion
%EndExpansion
%

%TCIMACRO{\FRAME{ftbpFU}{2.5884in}{2.5062in}{0pt}{\Qcb{Predictions of the
%overfitted model}}{\Qlb{f:predictions_checkerboard}}%
%{predictions_checkerboard.png}{\special{ language "Scientific Word";
%type "GRAPHIC";  maintain-aspect-ratio TRUE;  display "USEDEF";
%valid_file "F";  width 2.5884in;  height 2.5062in;  depth 0pt;
%original-width 4.7392in;  original-height 4.5887in;  cropleft "0";
%filename 'predictions_checkerboard.png';file-properties "XNPEU";}} }%
%BeginExpansion
%EndExpansion

\subsubsection{The polynomial regression with pairwise interactions}

The next numerical example uses the following regression function for
interpretation:
\begin{equation}
y(\mathbf{x})=x_{1}^{2}+x_{1}x_{2}-x_{3}x_{4}+x_{4}+0x_{5}+\varepsilon.
\end{equation}

\begin{figure}
[ptb]
\begin{center}
\includegraphics[
%%=6.987800in,
%%=9.098500in,
height=2.5305in,
width=3.2908in
]%
{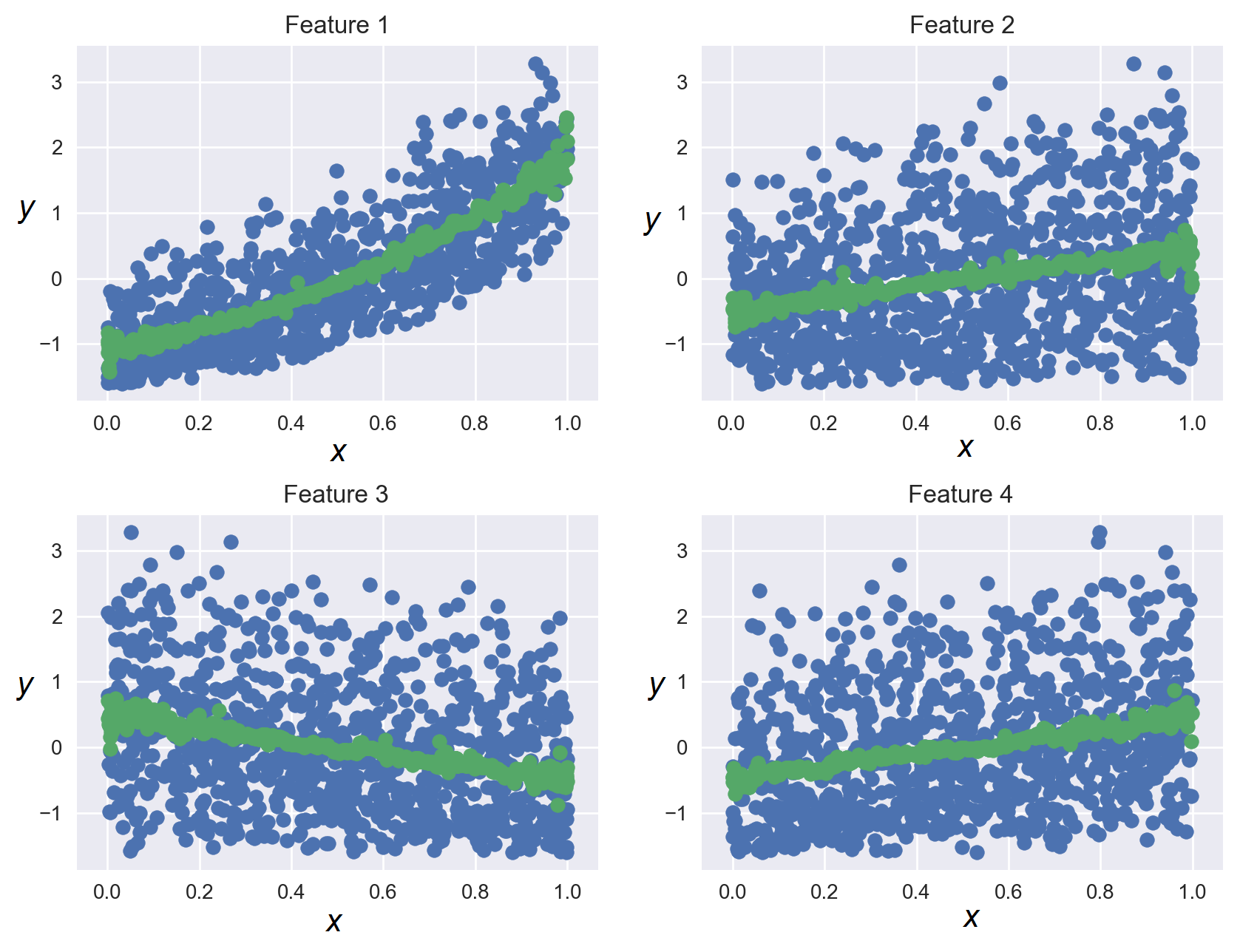}%
\caption{Predictions of single GBMs as functions of the corresponding feature
values}%
\label{f:featurewise_poly}%
\end{center}
\end{figure}

It contains pairwise interactions of features ($x_{1}x_{2}$ and $x_{3}x_{4}$).
Fig. \ref{f:history_nonlinear} shows how the weights are changed with increase
of the number of iterations $T$. The stabilization of weights is observed in
Fig. \ref{f:history_nonlinear}. This is an interesting fact because many
features are interacted. In spite of this interaction, the proposed algorithm
copes with this problem and provides the stabilized weights. Fig.
\ref{f:featurewise_poly} shows how predictions of each GBM depend on values of
the corresponding feature (see the explained similar Fig.
\ref{f:featurewise_linear}). One can see from Fig. \ref{f:featurewise_poly}
that the first feature significantly impacts on the prediction $y$ because the
narrow band rises significantly with the feature $x_{1}$.%

%TCIMACRO{\FRAME{ftbpFU}{3.6935in}{2.08in}{0pt}{\Qcb{Weights of features as
%functions of the iteration number for the polynomial regression}%
%}{\Qlb{f:history_poly}}{history_poly.png}%
%width 3.6935in;  height 2.08in;  depth 0pt;  original-width 9.2336in;
%original-height 5.1807in;  cropleft "0";  croptop "1";  cropright "1";
%filename 'history_poly.png';file-properties "XNPEU";}} }%
%BeginExpansion
\begin{figure}
[ptb]
\begin{center}
\includegraphics[
%%=5.180700in,
%%=9.233600in,
height=2.08in,
width=3.6935in
]%
{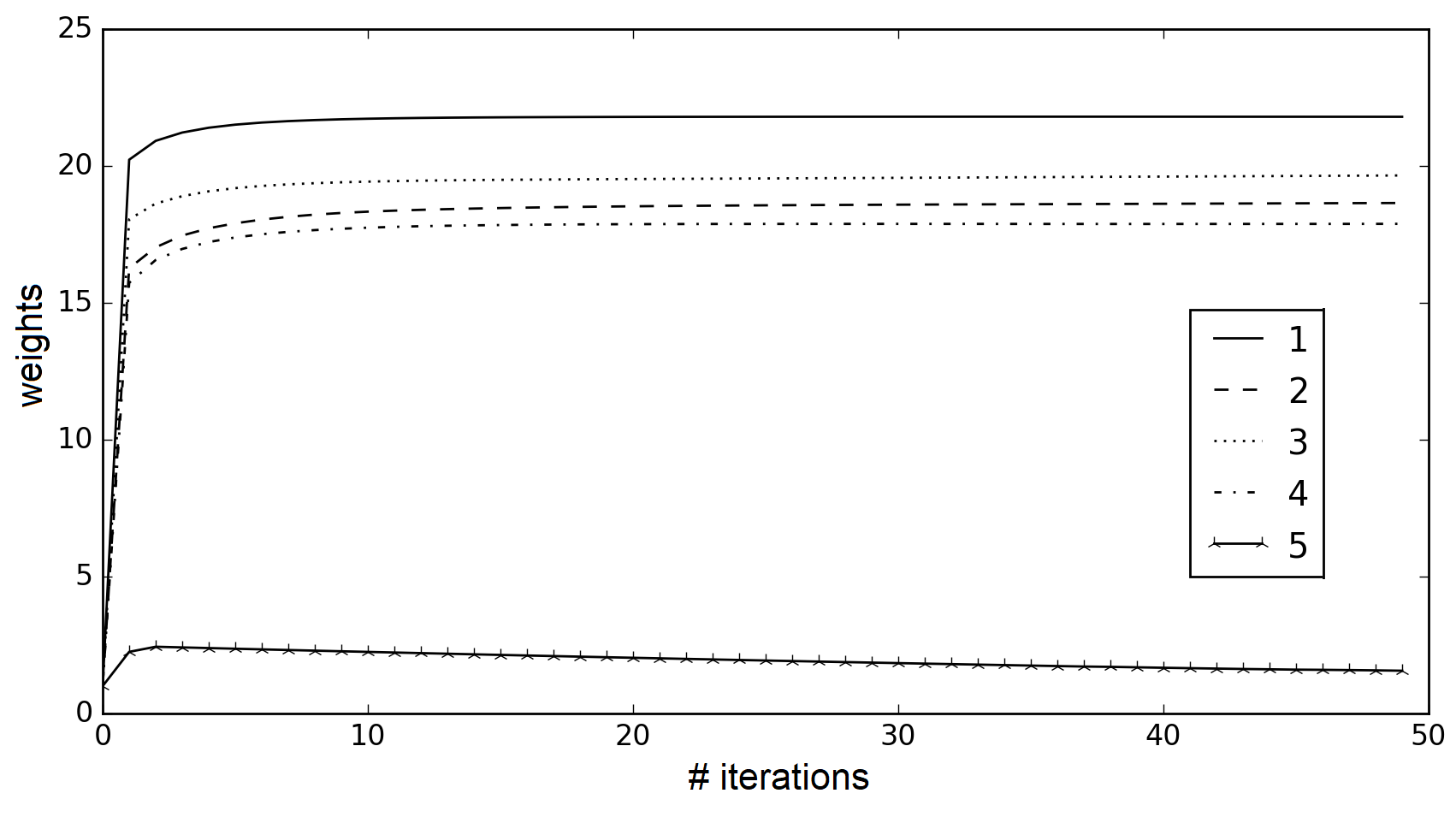}%
\caption{Weights of features as functions of the iteration number for the
polynomial regression}%
\label{f:history_poly}%
\end{center}
\end{figure}
%EndExpansion
%

%TCIMACRO{\FRAME{ftbpFU}{3.2908in}{2.5305in}{0pt}{\Qcb{Predictions of single
%GBMs as functions of the corresponding feature values}}%
%{\Qlb{f:featurewise_poly}}{featurewise_poly.png}%
%width 3.2908in;  height 2.5305in;  depth 0pt;  original-width 9.0985in;
%original-height 6.9878in;  cropleft "0";  croptop "1";  cropright "1";
%filename 'featurewise_poly.png';file-properties "XNPEU";}} }%
%BeginExpansion
%EndExpansion

The computed weights of features before their correction are
$(21.8,18.65,19.65,17.89,1.56)$. Weights after correction are
$(0.83,0.27,0.3,0.28,0.0)$. The weights imply that the first feature has the
highest importance. Indeed, it is presented in the polynomial twice: in
$x_{1}^{2}$ and in $x_{1}x_{2}$.

\subsection{Numerical experiments with real data}

\subsubsection{Boston Housing dataset}

Let us consider the real data called the Boston Housing dataset. It can be
obtained from the StatLib archive (http://lib.stat.cmu.edu/datasets/boston).
The Boston Housing dataset consists of 506 examples such that each example is
described by 13 features.%

%TCIMACRO{\FRAME{ftbpFU}{3.781in}{2.2303in}{0pt}{\Qcb{Importance of all
%features for the Boston Housing dataset in the case of the global
%interpretation}}{\Qlb{importances_boston}}{importances_boston.png}%
%{\special{ language "Scientific Word";  type "GRAPHIC";  display "USEDEF";
%valid_file "F";  width 3.781in;  height 2.2303in;  depth 0pt;
%original-width 10.4164in;  original-height 5.2087in;  cropleft "0";
%filename 'importances_boston.png';file-properties "XNPEU";}} }%
%BeginExpansion
\begin{figure}
[ptb]
\begin{center}
\includegraphics[
%%=5.208700in,
%%=10.416400in,
height=2.2303in,
width=3.781in
]%
{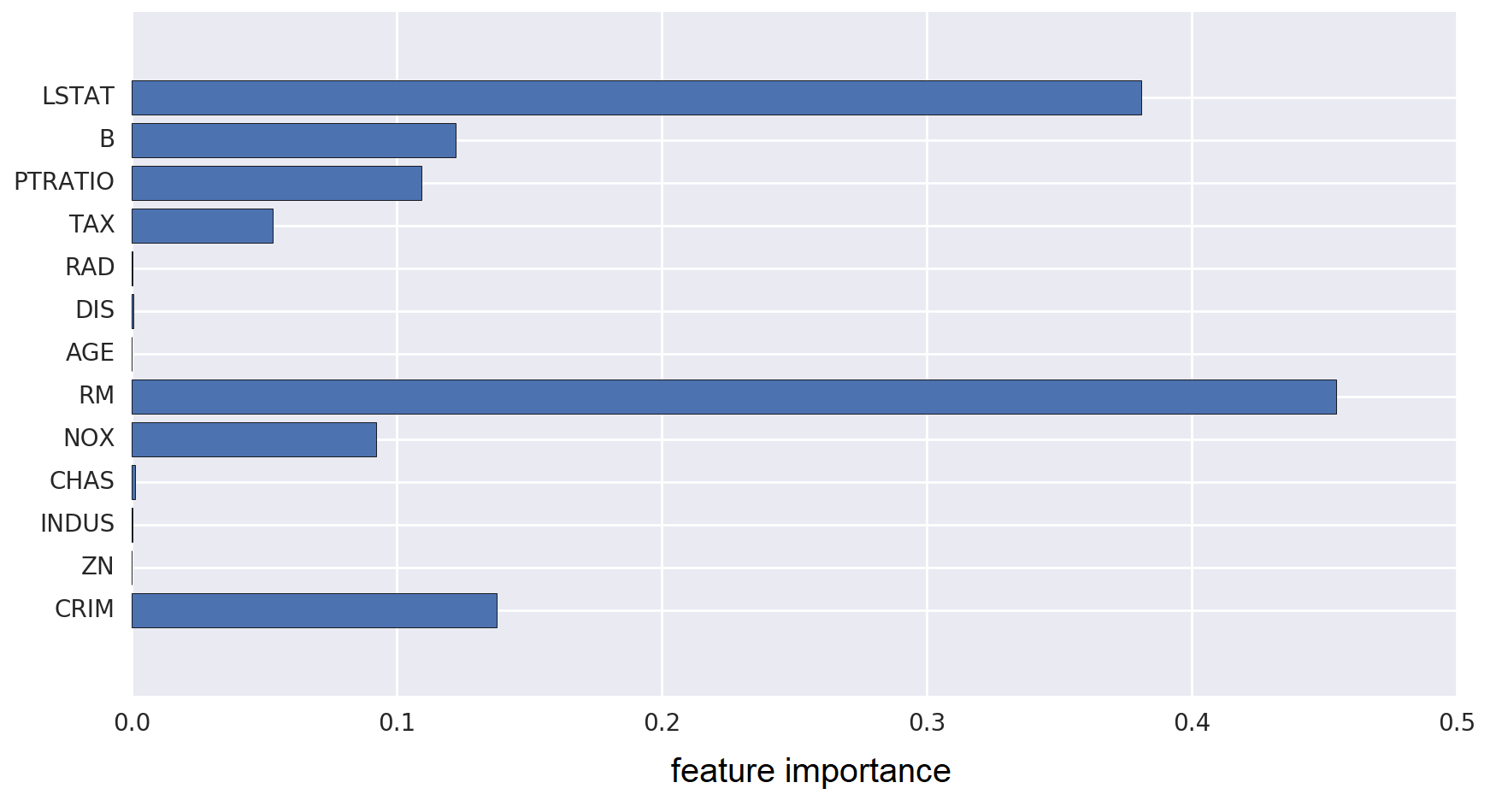}%
\caption{Importance of all features for the Boston Housing dataset in the case
of the global interpretation}%
\label{importances_boston}%
\end{center}
\end{figure}
%EndExpansion
%

%TCIMACRO{\FRAME{ftbpFU}{3.6019in}{2.1188in}{0pt}{\Qcb{Weights of the most
%important features as functions of the iteration number for the Boston Housing
%dataset and the global interpretation}}{\Qlb{f:history_boston}}%
%{history_boston.png}{\special{ language "Scientific Word";  type "GRAPHIC";
%width 3.6019in;  height 2.1188in;  depth 0pt;  original-width 9.193in;
%original-height 5.3956in;  cropleft "0";  croptop "1";  cropright "1";
%filename 'history_boston.png';file-properties "XNPEU";}} }%
%BeginExpansion
\begin{figure}
[ptb]
\begin{center}
\includegraphics[
%%=5.395600in,
%%=9.193000in,
height=2.1188in,
width=3.6019in
]%
{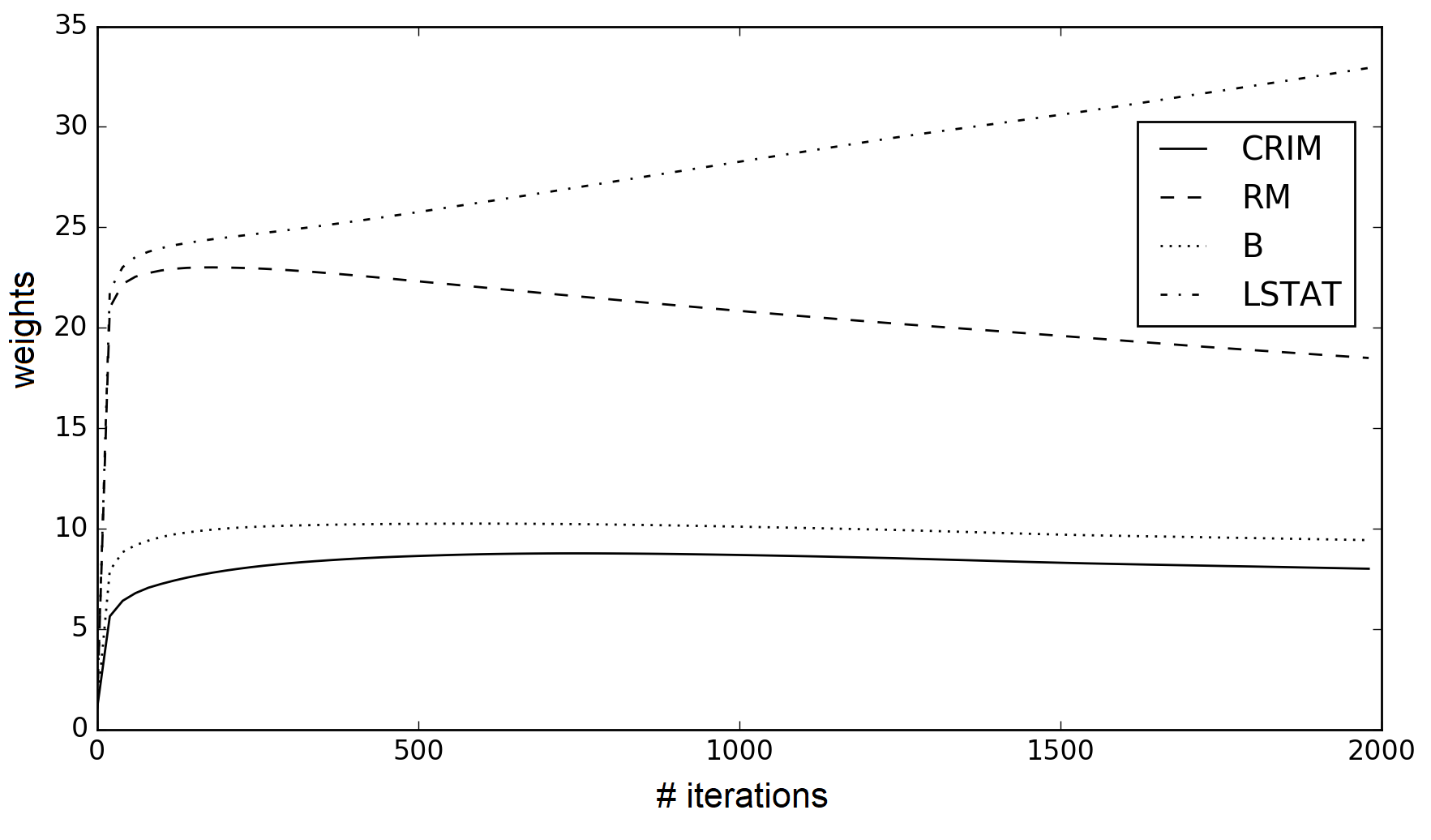}%
\caption{Weights of the most important features as functions of the iteration
number for the Boston Housing dataset and the global interpretation}%
\label{f:history_boston}%
\end{center}
\end{figure}
%EndExpansion

Importance of all features obtained by means of the proposed algorithm are
shown in Fig. \ref{importances_boston}. It can be seen from Fig.
\ref{importances_boston} that features CRIM, RM, B, LSTAT have the highest
importance values. Fig. \ref{f:history_boston} shows how weights are changed
with increase of the number of iterations $T$. Only four features having the
highest importance are shown in Fig. \ref{f:history_boston} in order to avoid
a confused mixture of many curves.%

%TCIMACRO{\FRAME{ftbpFU}{3.6079in}{2.789in}{0pt}{\Qcb{Shape functions of four
%important features learned on the Boston Housing dataset (the x-axis indicates
%values of each feature, the y-axis indicates the feature contriburion) for
%global interpretation}}{\Qlb{f:shape_boston}}{featurewise_boston.png}%
%width 3.6079in;  height 2.789in;  depth 0pt;  original-width 10.4167in;
%original-height 8.0419in;  cropleft "0";  croptop "1";  cropright "1";
%filename 'featurewise_boston.png';file-properties "XNPEU";}} }%
%BeginExpansion
\begin{figure}
[ptb]
\begin{center}
\includegraphics[
%%=8.041900in,
%%=10.416700in,
height=2.789in,
width=3.6079in
]%
{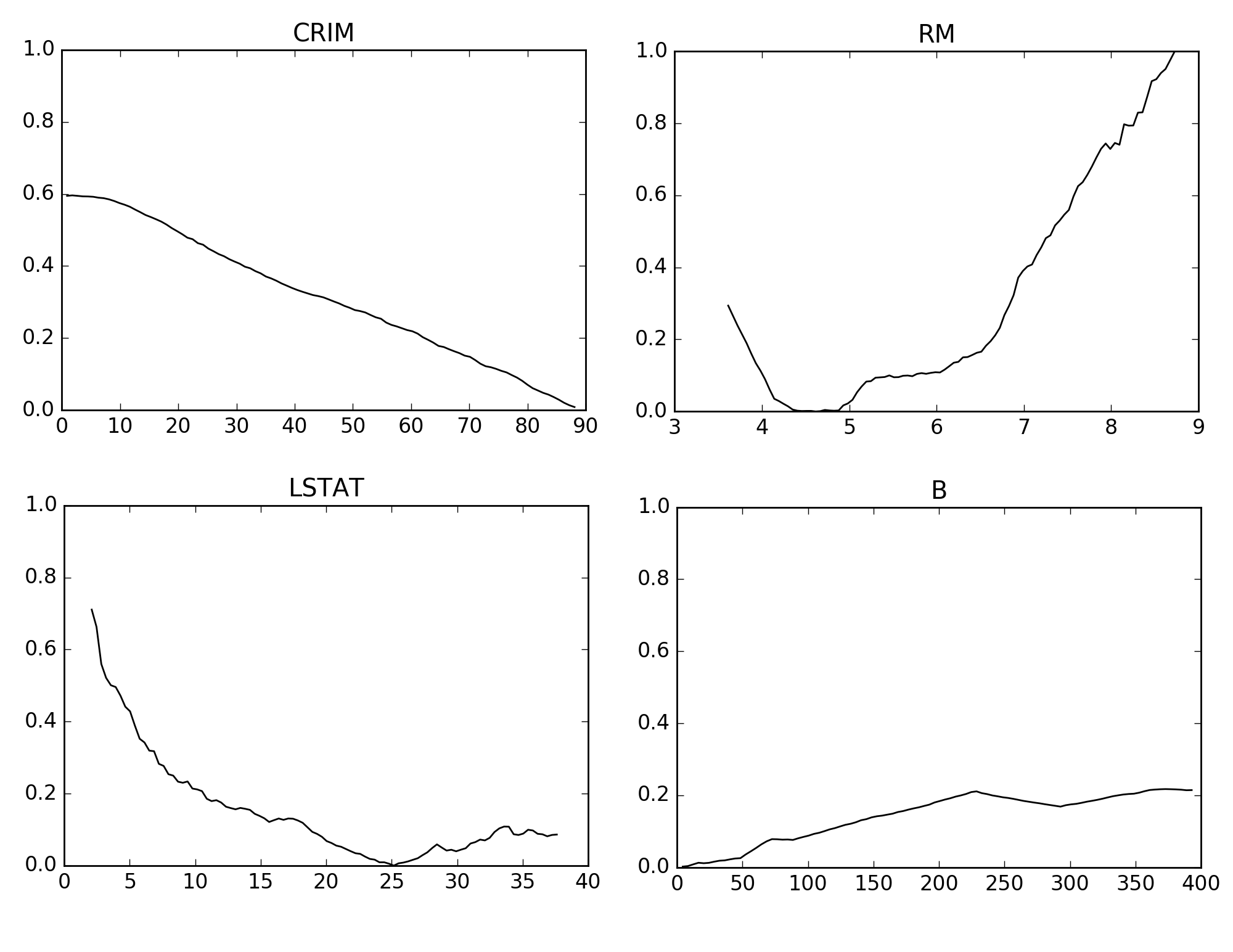}%
\caption{Shape functions of four important features learned on the Boston
Housing dataset (the x-axis indicates values of each feature, the y-axis
indicates the feature contriburion) for global interpretation}%
\label{f:shape_boston}%
\end{center}
\end{figure}
%EndExpansion

We plot individual shape functions $g_{k}(x_{k})$ also only for four important
features: RM, LSTAT, CRIM, B. They are depicted in Fig. \ref{f:shape_boston}.
Contributions of features ($y$-axis) are biased to 0 and scaled in order to
have the same interval from 0 to 1 for all plots. The shape plot for the most
important feature RM (the average number of rooms per dwelling) shows that
contribution of the RM rises significantly with the average number of rooms.
It is interesting to note that the contribution decreases when the number of
rooms is smaller than $4$. The shape plot for the second important feature
LSTAT (\% lower status of the population) shows that its contribution tends to
decrease. A small increase of the contribution when LSTAT is larger than $25$
may be caused by overfitting. Features CRIM and B have significantly smaller
weights. This fact can be seen from plots in Fig. \ref{f:shape_boston} where
changes of their contributions are small in comparison with the RM and the LSTAT.

\subsubsection{Breast Cancer dataset}

The next real dataset is the Breast Cancer Wisconsin (Diagnostic). It can be
found in the well-known UCI Machine Learning Repository
(https://archive.ics.uci.edu). The Breast Cancer dataset contains 569 examples
such that each example is described by 30 features. For classes of the breast
cancer diagnosis, the malignant and the benign are assigned by classes $0$ and
$1$, respectively. We consider the corresponding model in the framework of
regression with outcomes in the form of probabilities from 0 (malignant) to 1 (benign).

\begin{figure}
[ptb]
\begin{center}
\includegraphics[
%%=9.082300in,
%%=10.850600in,
height=3.2932in,
width=3.9306in
]%
{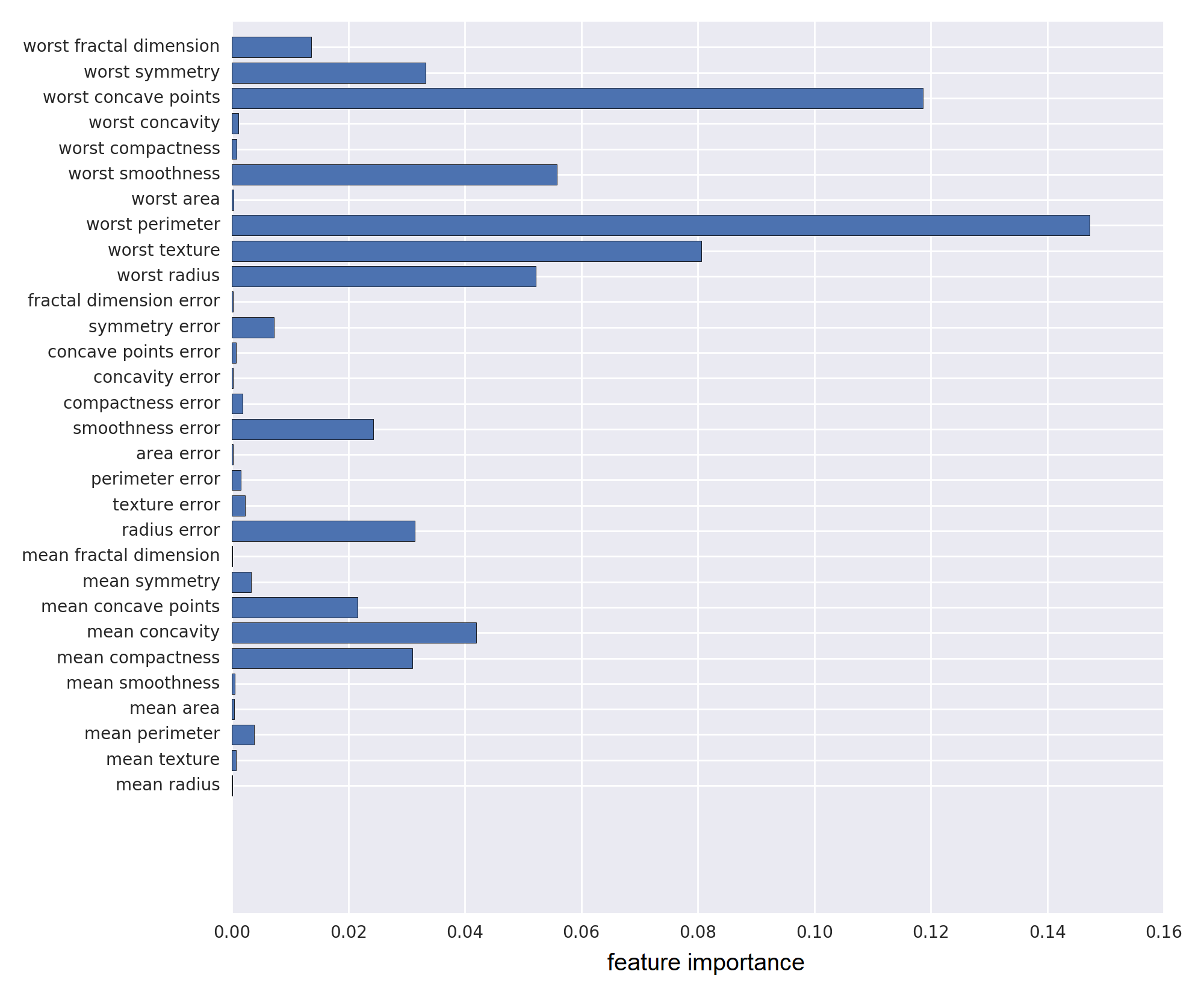}%
\caption{Importance of all features for the Breast Cancer dataset}%
\label{f:importances_bc}%
\end{center}
\end{figure}

\begin{figure}
[ptb]
\begin{center}
\includegraphics[
%%=5.115900in,
%%=9.396700in,
height=2.0418in,
width=3.7377in
]%
{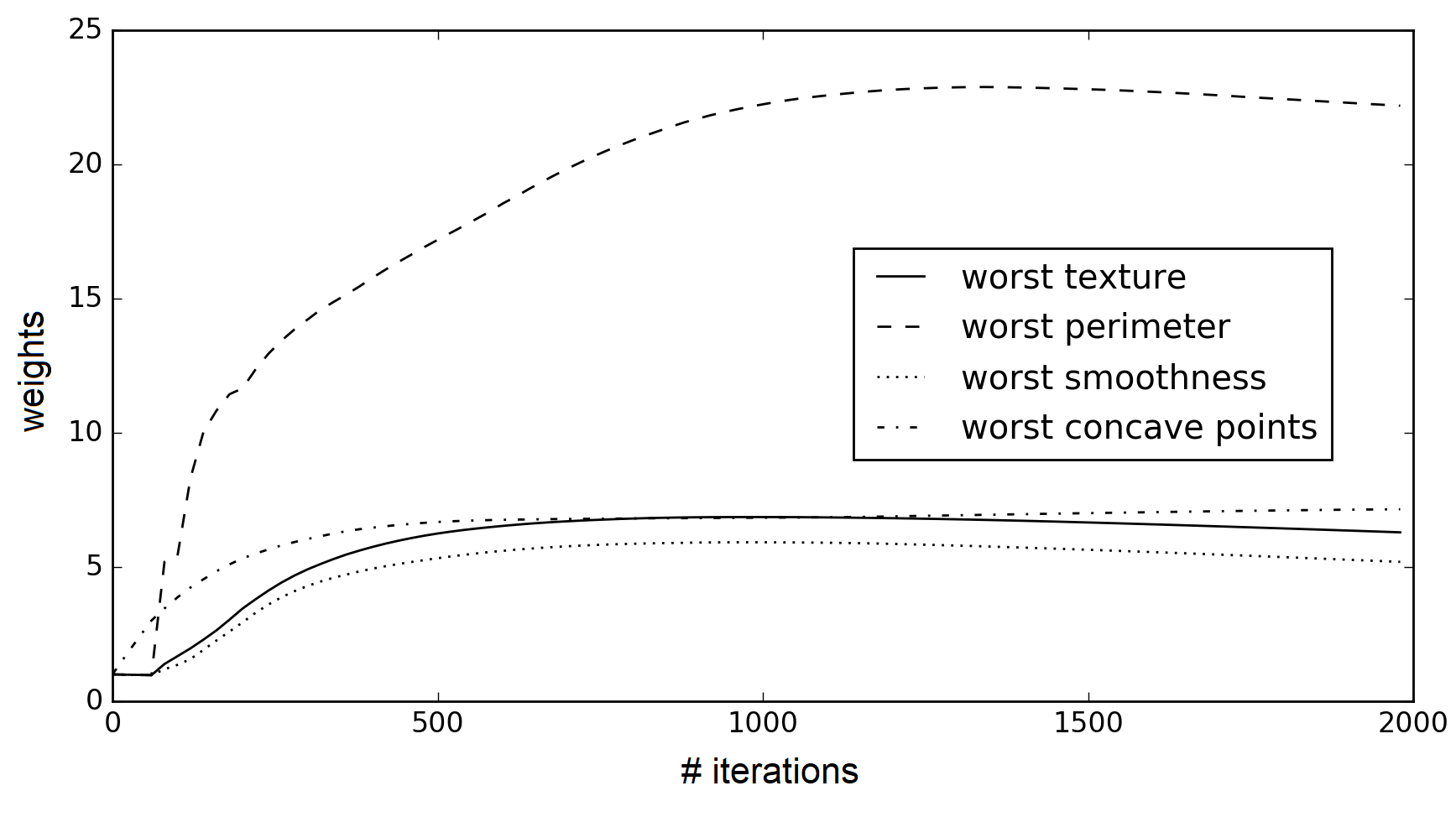}%
\caption{Weights of the most important features as functions of the iteration
number for the Breast Cancer dataset}%
\label{f:history_bc}%
\end{center}
\end{figure}

The importance values of features obtained by using the proposed algorithm are
depicted in Fig. \ref{f:importances_bc}. It can be seen from Fig.
\ref{f:importances_bc} that features \textquotedblleft worst
texture\textquotedblright, \textquotedblleft worst perimeter\textquotedblright%
, \textquotedblleft worst concave points\textquotedblright, \textquotedblleft
worst smoothness\textquotedblright\ are of the highest importance. Fig.
\ref{f:history_bc} shows how weights are changed with increase of the number
of iterations $T$. We again consider only four important features in Fig.
\ref{f:history_bc}.%

%TCIMACRO{\FRAME{ftbpFU}{3.9306in}{3.2932in}{0pt}{\Qcb{Importance of all
%features for the Breast Cancer dataset}}{\Qlb{f:importances_bc}}%
%{importances_bc.png}{\special{ language "Scientific Word";  type "GRAPHIC";
%width 3.9306in;  height 3.2932in;  depth 0pt;  original-width 10.8506in;
%original-height 9.0823in;  cropleft "0";  croptop "1";  cropright "1";
%filename 'importances_bc.png';file-properties "XNPEU";}} }%
%BeginExpansion
%EndExpansion
%

%TCIMACRO{\FRAME{ftbpFU}{3.7377in}{2.0418in}{0pt}{\Qcb{Weights of the most
%important features as functions of the iteration number for the Breast Cancer
%dataset}}{\Qlb{f:history_bc}}{history_bc.png}%
%width 3.7377in;  height 2.0418in;  depth 0pt;  original-width 9.3967in;
%original-height 5.1159in;  cropleft "0";  croptop "1";  cropright "1";
%cropbottom "0";  filename 'history_bc.png';file-properties "XNPEU";}}
%}%
%BeginExpansion
%EndExpansion

\begin{figure}
[ptb]
\begin{center}
\includegraphics[
%%=8.187200in,
%%=10.416700in,
height=3.026in,
width=3.845in
]%
{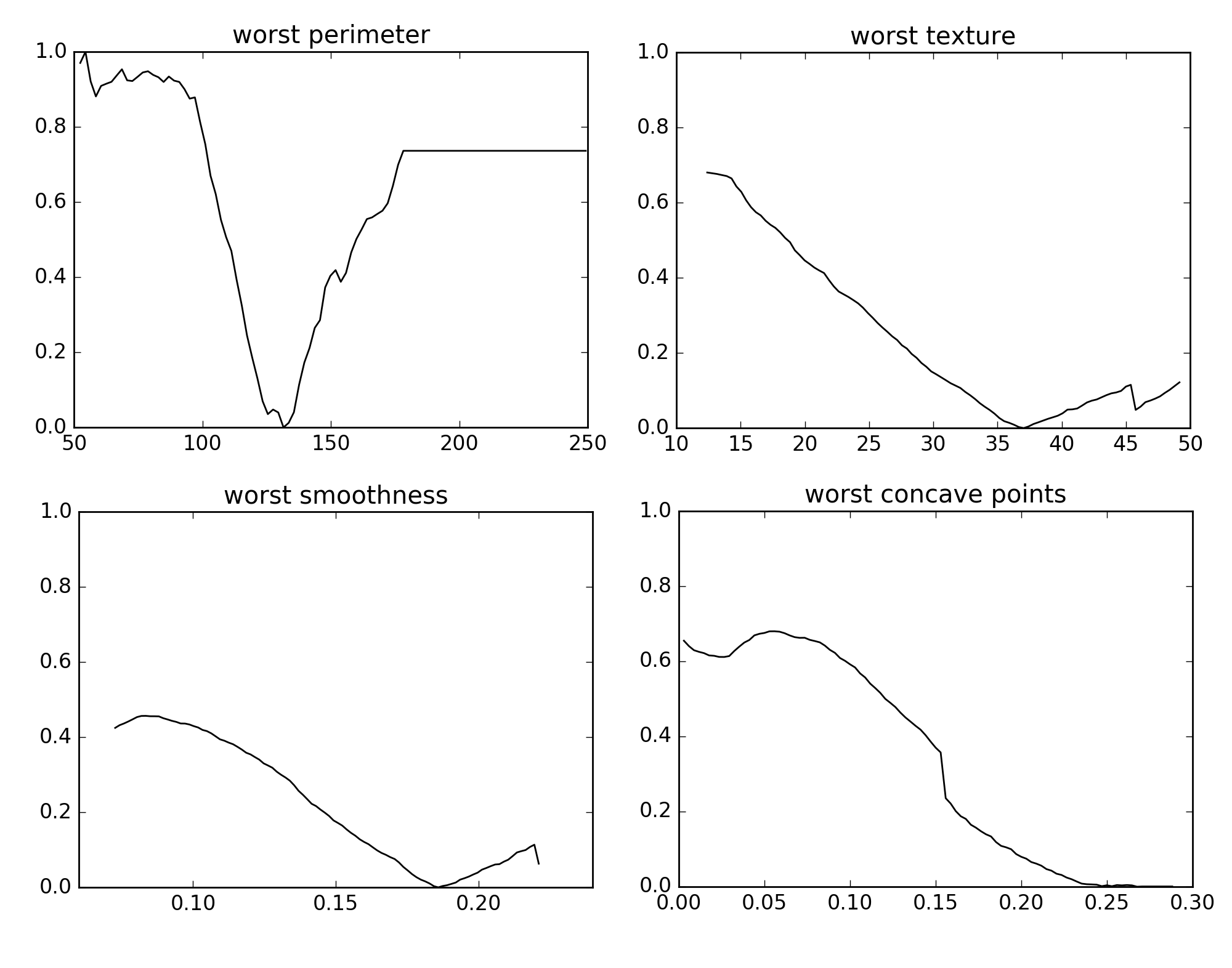}%
\caption{Shape functions of four important features learned on the Breast
Cancer dataset (the x-axis indicates values of each feature, the y-axis
indicates the feature contriburion) for the global interpretation}%
\label{f:curves_bc}%
\end{center}
\end{figure}

Individual shape functions are plotted for four important features:
\textquotedblleft worst texture\textquotedblright, \textquotedblleft worst
perimeter\textquotedblright, \textquotedblleft worst concave
points\textquotedblright, \textquotedblleft worst smoothness\textquotedblright%
. Fig. \ref{f:curves_bc} illustrates them. The shape plot for the
\textquotedblleft worst perimeter\textquotedblright\ shows that the
probability of benign drops with increase of the worst perimeter and increases
for the worst perimeter above 140. The shape plot for the second important
feature \textquotedblleft worst concave points\textquotedblright\ shows that
the probability of benign decreases with decrease of the worst concave points.
Features \textquotedblleft worst texture\textquotedblright\ and
\textquotedblleft worst smoothness\textquotedblright\ have significantly
smaller impact on the target probability.%

%TCIMACRO{\FRAME{ftbpFU}{3.845in}{3.026in}{0pt}{\Qcb{Shape functions of four
%important features learned on the Breast Cancer dataset (the x-axis indicates
%values of each feature, the y-axis indicates the feature contriburion) for the
%global interpretation}}{\Qlb{f:curves_bc}}{featurewise_bc.png}%
%width 3.845in;  height 3.026in;  depth 0pt;  original-width 10.4167in;
%original-height 8.1872in;  cropleft "0";  croptop "1";  cropright "1";
%filename 'featurewise_bc.png';file-properties "XNPEU";}} }%
%BeginExpansion
%EndExpansion

\section{Numerical experiments with the local interpretation}

So far, we have studied the global interpretation when tried to interpret all
examples in training sets. Let us consider numerical examples with the local interpretation.

\subsection{Numerical experiments with synthetic data}

\begin{figure}
[ptb]
\begin{center}
\includegraphics[
%%=4.651000in,
%%=9.312300in,
height=2.4344in,
width=4.8594in
]%
{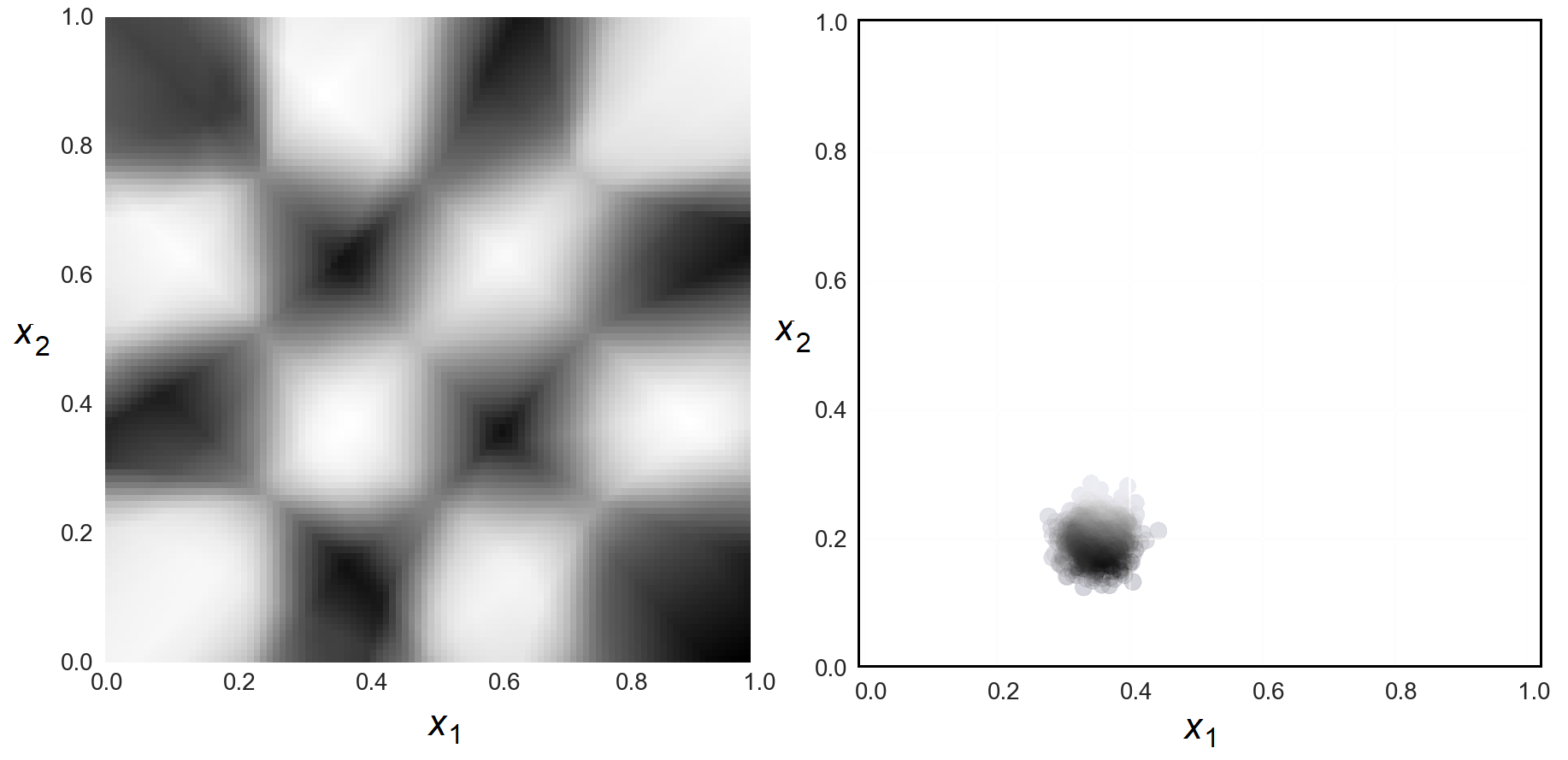}%
\caption{Predictions of the neural network learned on the generated
\textquotedblleft chess board\textquotedblright\ training set (left picture)
and perturbed points generated around point $(0.35,0.2)$ (right picture)}%
\label{f:pred_chess}%
\end{center}
\end{figure}

First, we return to the chess board example and investigate a bound between
two checkers. A neural network is used as a black-box regression model for
interpretation. The network is trained by using the quadratic loss function.
The black and white checkers are labelled by 1 and 0, respectively. The
network consists of 3 layers having 100 units with the ReLU activation
function and optimizer Adam. $N=1000$ perturbed points are generated around
point $(0.35,0.2)$ from the normal distribution with the standard deviation
$0.025$. The perturbed points are depicted in Fig. \ref{f:pred_chess} (the
right picture). Predictions of the neural network learned on the generated
\textquotedblleft chess board\textquotedblright\ training set are depicted in
Fig. \ref{f:pred_chess} (the left picture).%

%TCIMACRO{\FRAME{ftbpFU}{4.8594in}{2.4344in}{0pt}{\Qcb{Predictions of the
%neural network learned on the generated \textquotedblleft chess
%board\textquotedblright\ training set (left picture) and perturbed points
%generated around point $(0.35,0.2)$ (right picture)}}{\Qlb{f:pred_chess}%
%}{predictions_mlp_checkerboard.png}{\special{ language "Scientific Word";
%type "GRAPHIC";  maintain-aspect-ratio TRUE;  display "USEDEF";
%valid_file "F";  width 4.8594in;  height 2.4344in;  depth 0pt;
%original-width 9.3123in;  original-height 4.651in;  cropleft "0";
%filename 'predictions_mlp_checkerboard.png';file-properties "XNPEU";}%
%} }%
%BeginExpansion
%EndExpansion

\begin{figure}
[ptb]
\begin{center}
\includegraphics[
%%=5.411100in,
%%=5.364400in,
height=2.2805in,
width=2.2615in
]%
{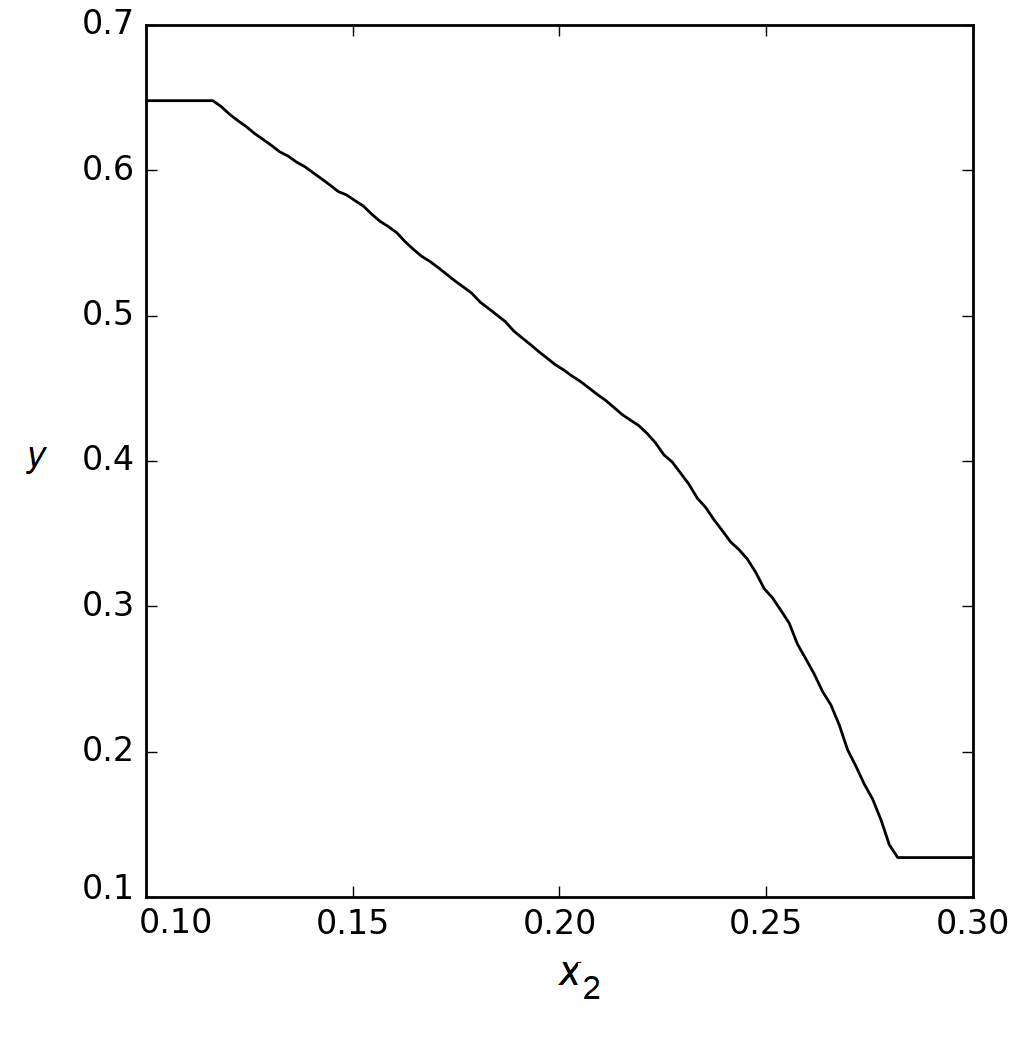}%
\caption{The shape function of the second feature $x_{2}$ for the chess board
example (the x-axis indicates values of feature $x_{2}$, the y-axis indicates
the probability of the black or white checkers)}%
\label{f:expl_chess_x2}%
\end{center}
\end{figure}

Fig. \ref{f:expl_chess_x2} depicts the shape function of the second feature
$x_{2}$ for the chess board example. It follows from Fig.
\ref{f:expl_chess_x2} that the probability of the class $1$ (the black
checker) decreases with increase of the second feature. This is also seen from
Fig. \ref{f:pred_chess}, where generated points \textquotedblleft
move\textquotedblright\ from the black checker to the white checker.
%TCIMACRO{\FRAME{ftbpFU}{2.2615in}{2.2805in}{0pt}{\Qcb{The shape function of
%the second feature $x_{2}$ for the chess board example (the x-axis indicates
%values of feature $x_{2}$, the y-axis indicates the probability of the black
%or white checkers)}}{\Qlb{f:expl_chess_x2}}%
%{explanation_checkerboard_nlp_y.png}{\special{ language "Scientific Word";
%type "GRAPHIC";  maintain-aspect-ratio TRUE;  display "USEDEF";
%valid_file "F";  width 2.2615in;  height 2.2805in;  depth 0pt;
%original-width 5.3644in;  original-height 5.4111in;  cropleft "0";
%filename 'explanation_checkerboard_nlp_y.png';file-properties "XNPEU";}%
%} }%
%BeginExpansion
%EndExpansion

\subsection{Numerical experiments with real data}

\subsubsection{Boston Housing dataset}

Let us consider the Boston Housing dataset for the local interpretation. A
point of interest is randomly selected by means of the following procedure.
First, two points $\mathbf{x}_{1}$ and $\mathbf{x}_{2}$ are randomly
selected\textbf{ }from the dataset. Then the point for interpretation is
determined as the middle point between $\mathbf{x}_{1}$ and $\mathbf{x}_{2}%
$\textbf{.} The neural network consisting of 3 layers having 100 units with
the ReLU activation function and optimizer Adam is used for training on the
Boston Housing dataset. The learning rate is $10^{-3}$, the number of epochs
is $1000$. $N=1000$ perturbed points are generated around the point of
interest from the uniform distribution with bounds $\mathbf{x}_{1}$ and
$\mathbf{x}_{2}$. All features are standardized before training the neural
network and interpreting results.

\begin{figure}
[ptb]
\begin{center}
\includegraphics[
%%=4.937200in,
%%=9.077900in,
height=2.0531in,
width=3.7645in
]%
{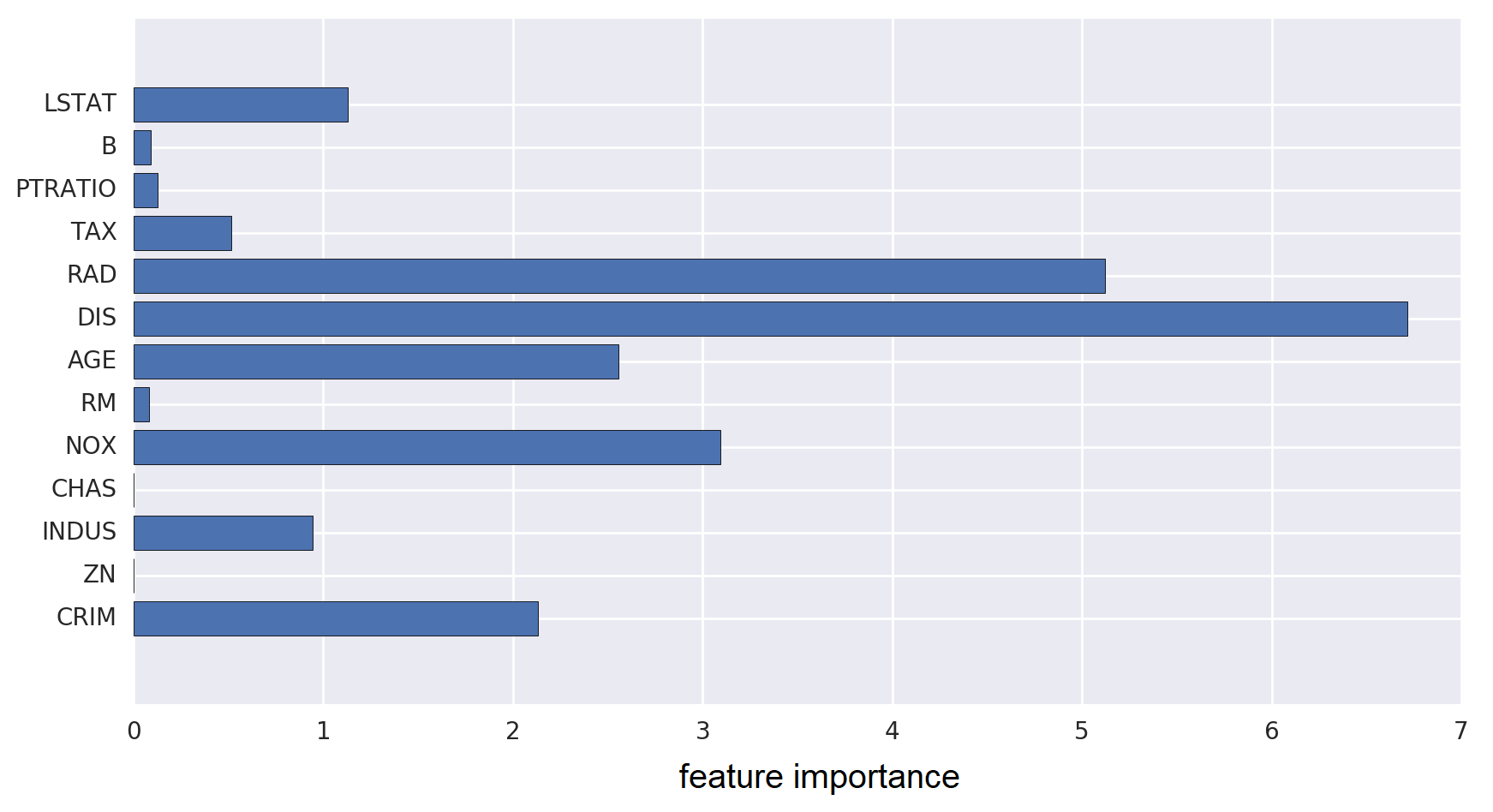}%
\caption{Importance of features for the Boston Housing dataset by considering
the local interpretation}%
\label{f:importances_boston_loc}%
\end{center}
\end{figure}

\begin{figure}
[ptb]
\begin{center}
\includegraphics[
%%=5.093800in,
%%=9.046800in,
height=2.0141in,
width=3.5674in
]%
{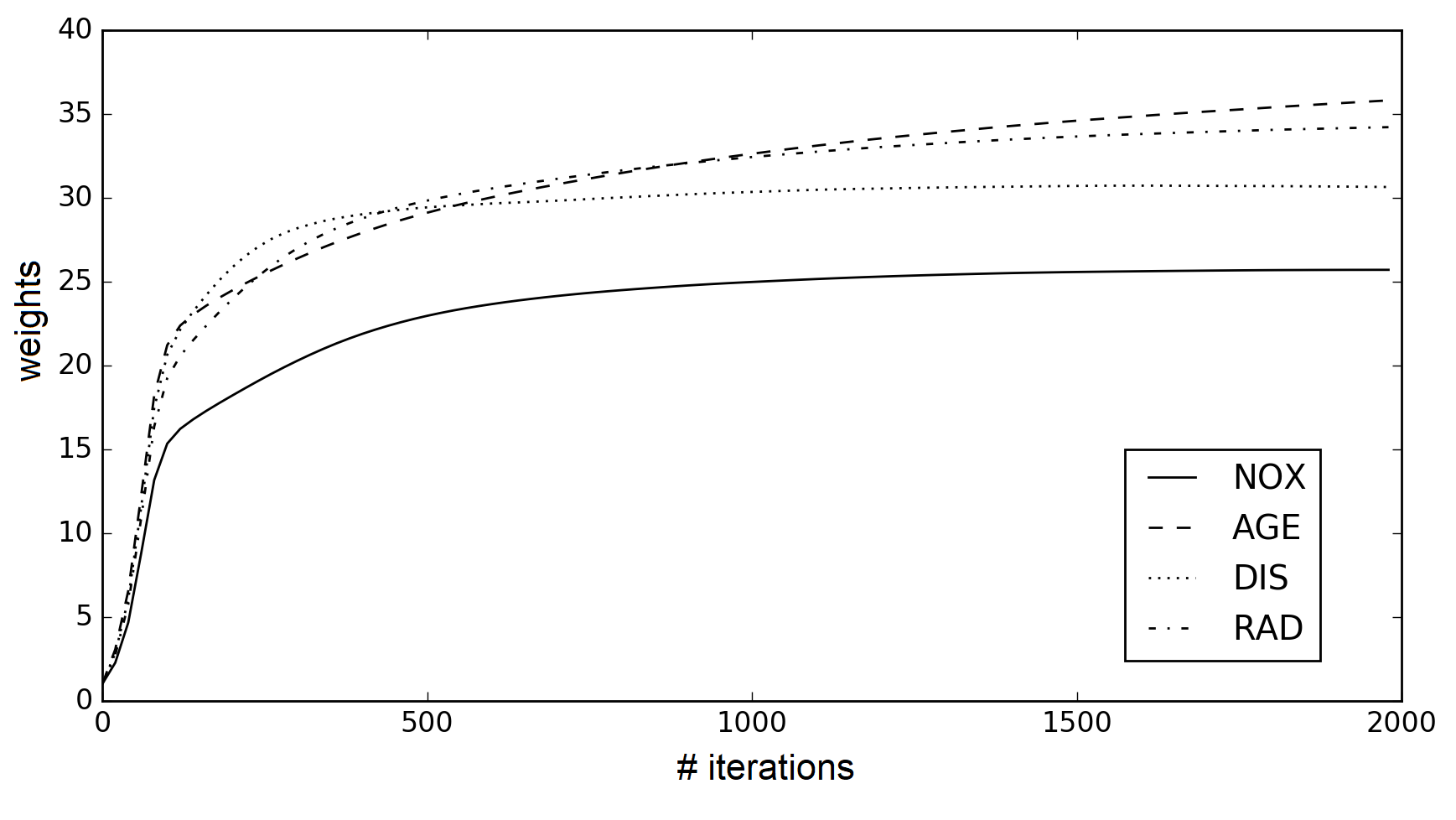}%
\caption{Weights of the most important features as functions of the iteration
number for the Boston Housing dataset and the local interpretation}%
\label{f:history_boston_loc}%
\end{center}
\end{figure}

Importance values of all features obtained by means of the proposed method are
shown in Fig. \ref{f:importances_boston_loc}. It follows from Fig.
\ref{f:importances_boston_loc} that features NOX, AGE, DIS, RAD have the
highest importance. They differ from the important features obtained for the
case of global interpretation (see Fig. \ref{importances_boston}). Fig.
\ref{f:history_boston_loc} shows how the weights are changed with increase of
the number of iterations $T$. Only four important features are shown in Fig.
\ref{f:history_boston_loc}.%

%TCIMACRO{\FRAME{ftbpFU}{3.7645in}{2.0531in}{0pt}{\Qcb{Importance of features
%for the Boston Housing dataset by considering the local interpretation}%
%}{\Qlb{f:importances_boston_loc}}{importances_boston_loc.png}%
%width 3.7645in;  height 2.0531in;  depth 0pt;  original-width 9.0779in;
%original-height 4.9372in;  cropleft "0";  croptop "1";  cropright "1";
%filename 'importances_boston_loc.png';file-properties "XNPEU";}} }%
%BeginExpansion
%EndExpansion
%

%TCIMACRO{\FRAME{ftbpFU}{3.5674in}{2.0141in}{0pt}{\Qcb{Weights of the most
%important features as functions of the iteration number for the Boston Housing
%dataset and the local interpretation}}{\Qlb{f:history_boston_loc}%
%}{history_boston_loc.png}{\special{ language "Scientific Word";
%type "GRAPHIC";  maintain-aspect-ratio TRUE;  display "USEDEF";
%valid_file "F";  width 3.5674in;  height 2.0141in;  depth 0pt;
%original-width 9.0468in;  original-height 5.0938in;  cropleft "0";
%filename 'history_boston_loc.png';file-properties "XNPEU";}} }%
%BeginExpansion
%EndExpansion
%

%TCIMACRO{\FRAME{ftbpFU}{3.6478in}{2.7614in}{0pt}{\Qcb{Shape functions of four
%important features learned on the Boston Housing dataset (the x-axis indicates
%values of each feature, the y-axis indicates the feature contriburion) for
%local interpretation}}{\Qlb{f:featurewise_boston_loc}}%
%{featurewise_boston_loc.png}{\special{ language "Scientific Word";
%type "GRAPHIC";  maintain-aspect-ratio TRUE;  display "USEDEF";
%valid_file "F";  width 3.6478in;  height 2.7614in;  depth 0pt;
%original-width 10.4167in;  original-height 7.875in;  cropleft "0";
%filename 'featurewise_boston_loc.png';file-properties "XNPEU";}} }%
%BeginExpansion
\begin{figure}
[ptb]
\begin{center}
\includegraphics[
%%=7.875000in,
%%=10.416700in,
height=2.7614in,
width=3.6478in
]%
{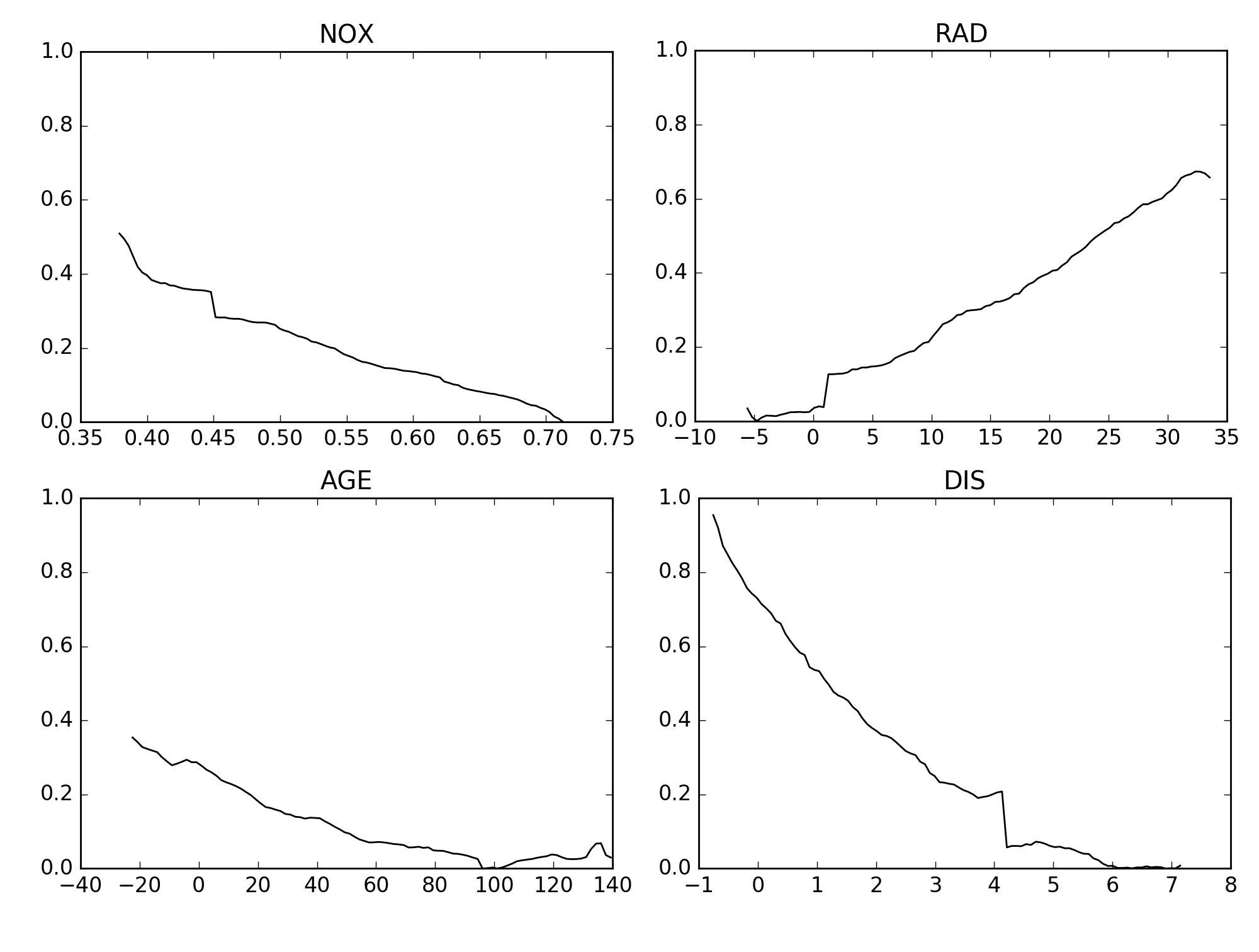}%
\caption{Shape functions of four important features learned on the Boston
Housing dataset (the x-axis indicates values of each feature, the y-axis
indicates the feature contriburion) for local interpretation}%
\label{f:featurewise_boston_loc}%
\end{center}
\end{figure}
%EndExpansion

Shape functions are depicted for four important features NOX, AGE, DIS, RAD in
Fig. \ref{f:featurewise_boston_loc}. The shape plot for the most important
feature RM shows that contribution of the RM rises significantly with the
average number of rooms. It is interesting to note that the contribution
decreases when the number of rooms is smaller than $4$. The shape plot for the
second important feature LSTAT shows that its contribution tends to decrease.
A small increase of the contribution, when LSTAT is larger than $25$, may be
caused by overfitting. Features CRIM and B have significantly smaller weights.
This fact can be seen from plots in Fig. \ref{f:featurewise_boston_loc} where
changes of their contributions are small in comparison with the RM and LSTAT.

\subsubsection{Breast Cancer dataset and the regression black-box model}

For illustrating the local interpretation, we again use the Breast Cancer
Wisconsin (Diagnostic) dataset. For classes of the breast cancer diagnosis,
the malignant and the benign are assigned by labels $0$ and $1$, respectively.
We consider the corresponding model in the framework of regression with
outcomes in the form of probabilities from 0 (malignant) to 1 (benign), i.e.,
we use the regression black-box model based on the SVM with the RBF kernel
having parameter $\gamma=1/m$. The penalty parameter $C$ of the SVM is $1.0$.
The point of interest is determined by means of the same procedure as for the
Boston Housing dataset. The perturbation procedure also coincides with the
same procedure for the Boston Housing dataset. All features are standardized
before training the SVM and interpreting results.

\begin{figure}
[ptb]
\begin{center}
\includegraphics[
%%=8.593600in,
%%=10.416700in,
height=3.7567in,
width=4.5489in
]%
{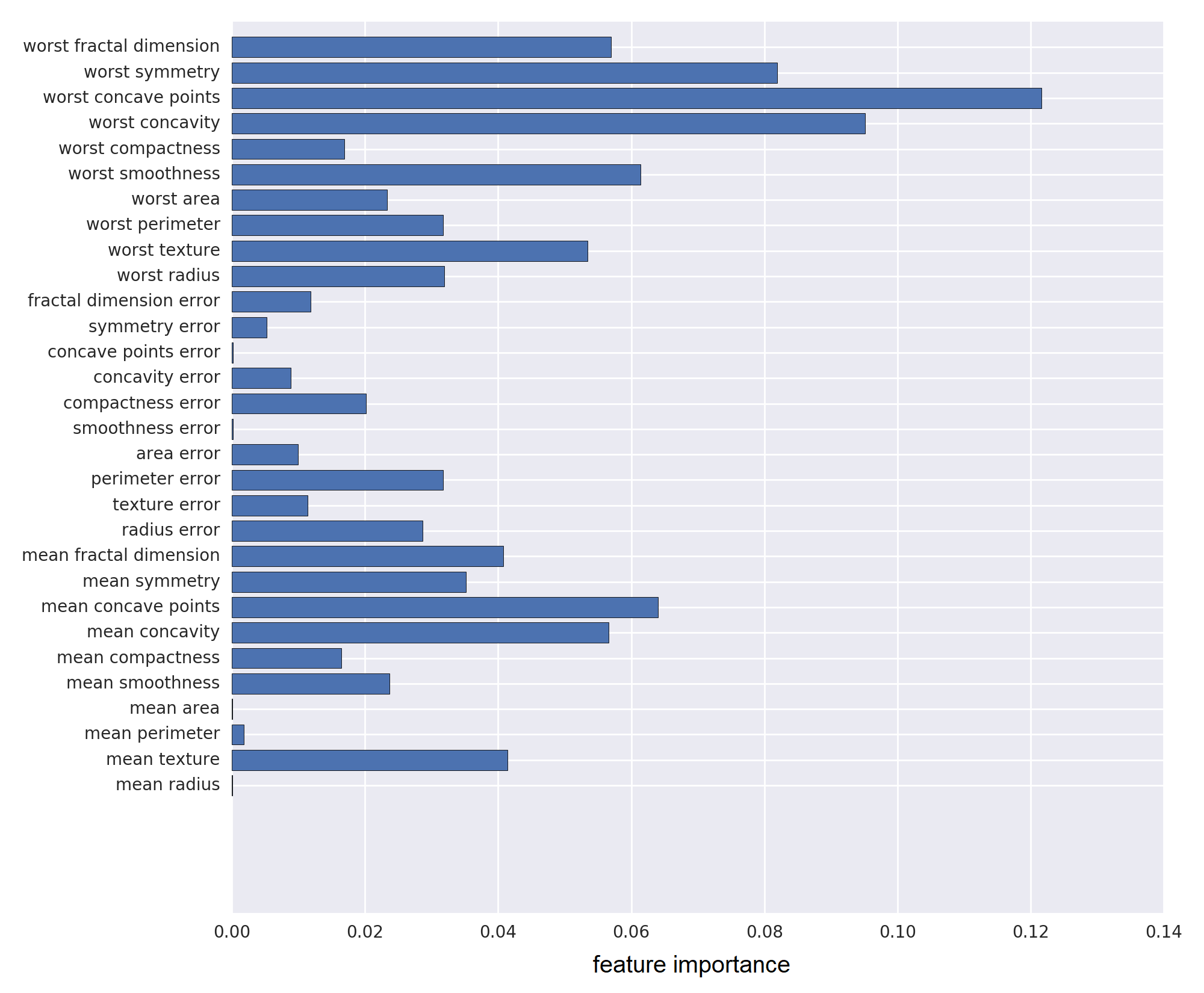}%
\caption{Importance of all features for the Breast Cancer dataset in the case
of the local interpretation using the regression black-box model}%
\label{f:importances_bc_loc}%
\end{center}
\end{figure}

\begin{figure}
[ptb]
\begin{center}
\includegraphics[
%%=5.104100in,
%%=9.151400in,
height=2.0539in,
width=3.6729in
]%
{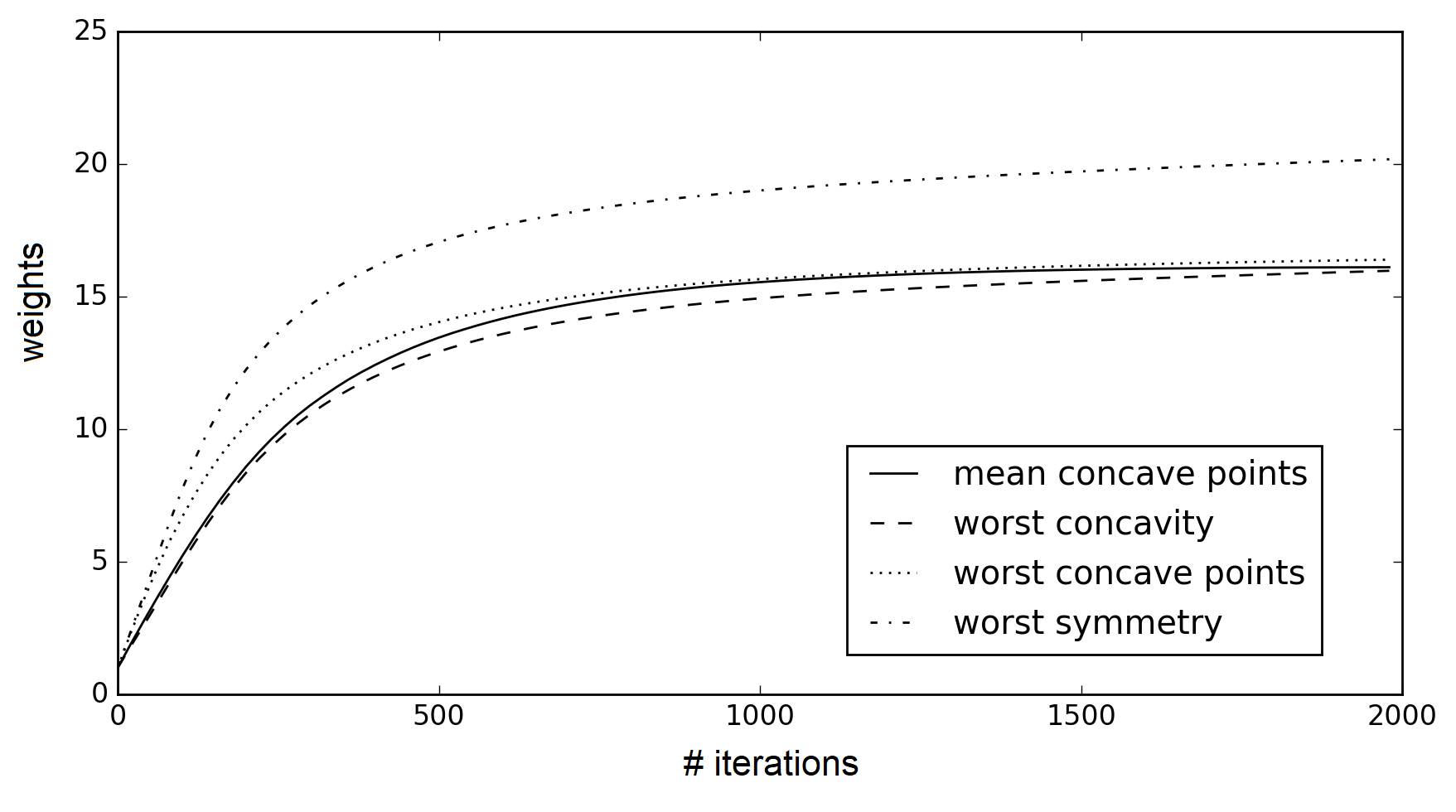}%
\caption{Weights of the most important features as functions of the iteration
number for the Breast Cancer dataset and the local interpretation using the
regression black-box model}%
\label{f:history_bc_loc}%
\end{center}
\end{figure}

Importance of features obtained by using the proposed algorithm are depicted
in Fig. \ref{f:importances_bc_loc}. It can be seen from Fig.
\ref{f:importances_bc_loc} that features \textquotedblleft worst
symmetry\textquotedblright, \textquotedblleft mean concave
points\textquotedblright, \textquotedblleft worst concavity\textquotedblright,
\textquotedblleft worst concave points\textquotedblright\ are of the highest
importance. They mainly differ from the important features obtained for the
global interpretation. Fig. \ref{f:history_bc_loc} shows how the weights are
changed with increase of the number of iterations $T$ for the above important features.%

%TCIMACRO{\FRAME{ftbpFU}{4.5489in}{3.7567in}{0pt}{\Qcb{Importance of all
%features for the Breast Cancer dataset in the case of the local interpretation
%using the regression black-box model}}{\Qlb{f:importances_bc_loc}%
%}{importances_bc_loc.png}{\special{ language "Scientific Word";
%type "GRAPHIC";  maintain-aspect-ratio TRUE;  display "USEDEF";
%valid_file "F";  width 4.5489in;  height 3.7567in;  depth 0pt;
%original-width 10.4167in;  original-height 8.5936in;  cropleft "0";
%filename 'importances_bc_loc.png';file-properties "XNPEU";}} }%
%BeginExpansion
%EndExpansion
%

%TCIMACRO{\FRAME{ftbpFU}{3.6729in}{2.0539in}{0pt}{\Qcb{Weights of the most
%important features as functions of the iteration number for the Breast Cancer
%dataset and the local interpretation using the regression black-box model}%
%}{\Qlb{f:history_bc_loc}}{history_bc_loc.png}%
%width 3.6729in;  height 2.0539in;  depth 0pt;  original-width 9.1514in;
%original-height 5.1041in;  cropleft "0";  croptop "1";  cropright "1";
%filename 'history_bc_loc.png';file-properties "XNPEU";}} }%
%BeginExpansion
%EndExpansion

\begin{figure}
[ptb]
\begin{center}
\includegraphics[
%%=7.771200in,
%%=10.416700in,
height=2.783in,
width=3.7256in
]%
{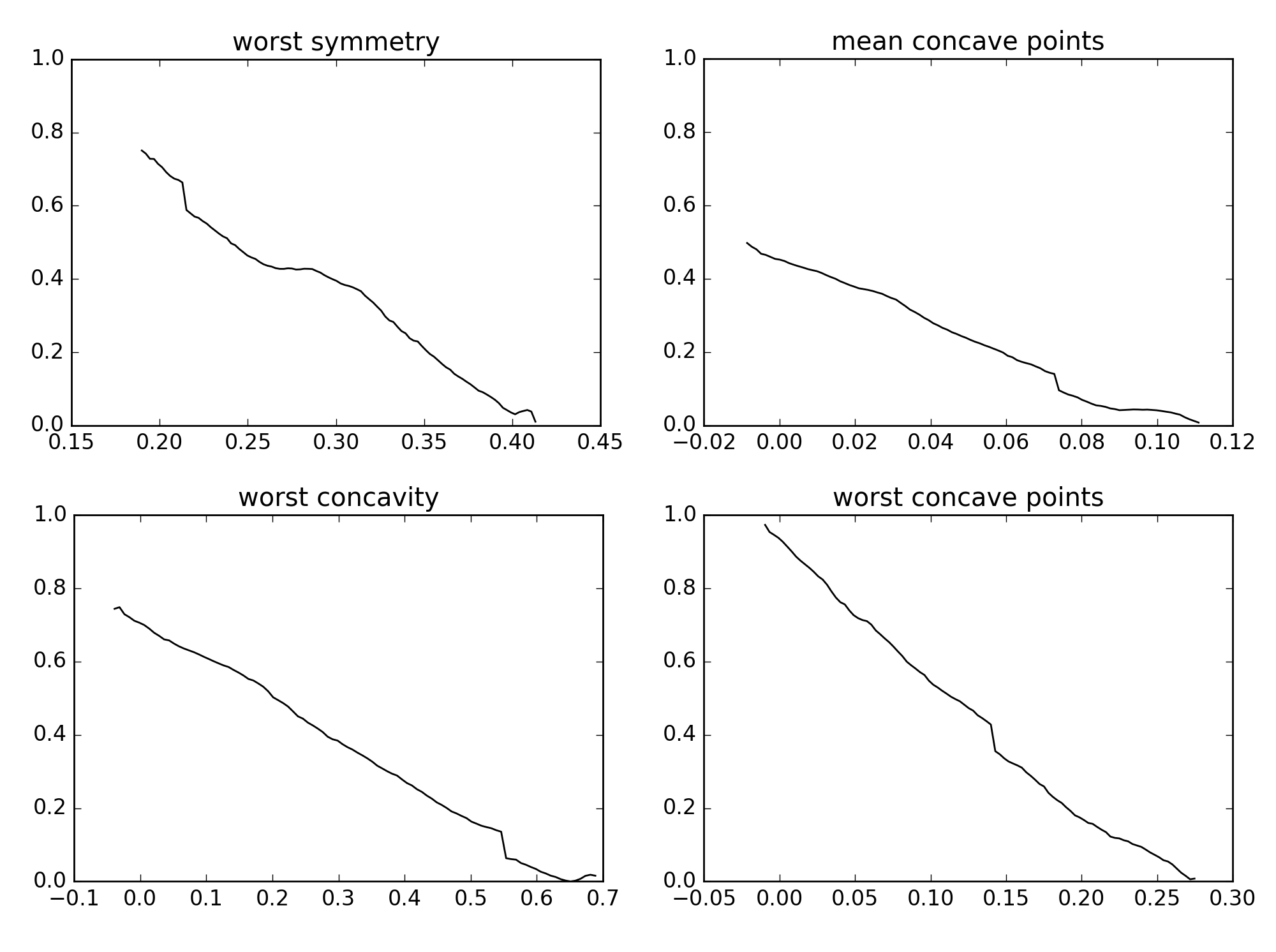}%
\caption{Shape functions of four important features learned on the Breast
Cancer dataset (the x-axis indicates values of each feature, the y-axis
indicates the feature contriburion) for the local interpretation using the
regression black-box model}%
\label{f:featurewise_bc_loc}%
\end{center}
\end{figure}

For these important features, individual shape functions are plotted in Fig.
\ref{f:featurewise_bc_loc}. The shape plot for the \textquotedblleft worst
concave points\textquotedblright\ shows that the probability of benign drops
with increase of the worst concave points. The shape plot for the second
important feature \textquotedblleft worst concavity\textquotedblright%
\ also\ shows that the probability of benign significantly decreases with
increase of the worst concavity. Features \textquotedblleft worst
symmetry\textquotedblright\ and \textquotedblleft mean concave
points\textquotedblright\ have significantly smaller impacts on the target probability.%

%TCIMACRO{\FRAME{ftbpFU}{3.7256in}{2.783in}{0pt}{\Qcb{Shape functions of four
%important features learned on the Breast Cancer dataset (the x-axis indicates
%values of each feature, the y-axis indicates the feature contriburion) for the
%local interpretation using the regression black-box model}}%
%{\Qlb{f:featurewise_bc_loc}}{featurewise_bc_loc.png}%
%width 3.7256in;  height 2.783in;  depth 0pt;  original-width 10.4167in;
%original-height 7.7712in;  cropleft "0";  croptop "1";  cropright "1";
%filename 'featurewise_bc_loc.png';file-properties "XNPEU";}} }%
%BeginExpansion
%EndExpansion

\subsubsection{Breast Cancer dataset and the classification black-box model}

We consider the Breast Cancer Wisconsin (Diagnostic) dataset and corresponding
model in the framework of classification with the same outcomes in the form of
probabilities from 0 (malignant) to 1 (benign), i.e., we use the
classification black-box model based on the SVM. Parameters of the black-box
model, the point of interest and the perturbation procedure do not differ from
those used for the Breast Cancer dataset with the regression black-box model.
All features are standardized before training the SVM and interpreting results.

\begin{figure}
[ptb]
\begin{center}
\includegraphics[
%%=9.038100in,
%%=10.850600in,
height=3.4039in,
width=4.0811in
]%
{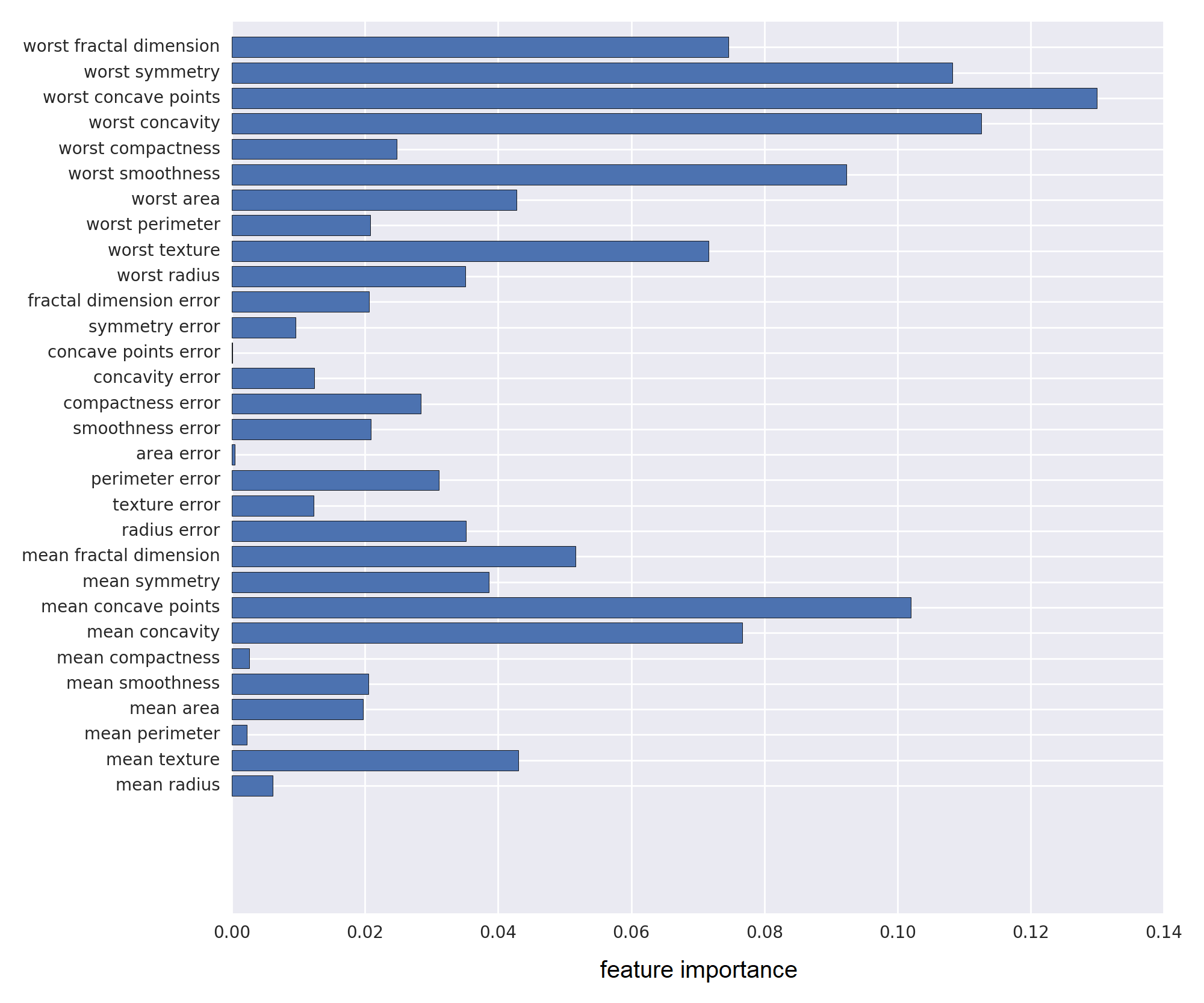}%
\caption{Importance of all features for the Breast Cancer dataset in the case
of the local interpretation using the classification black-box model}%
\label{f:importances_bc_clf_loc}%
\end{center}
\end{figure}

\begin{figure}
[ptb]
\begin{center}
\includegraphics[
%%=5.306000in,
%%=9.477700in,
height=2.1188in,
width=3.7727in
]%
{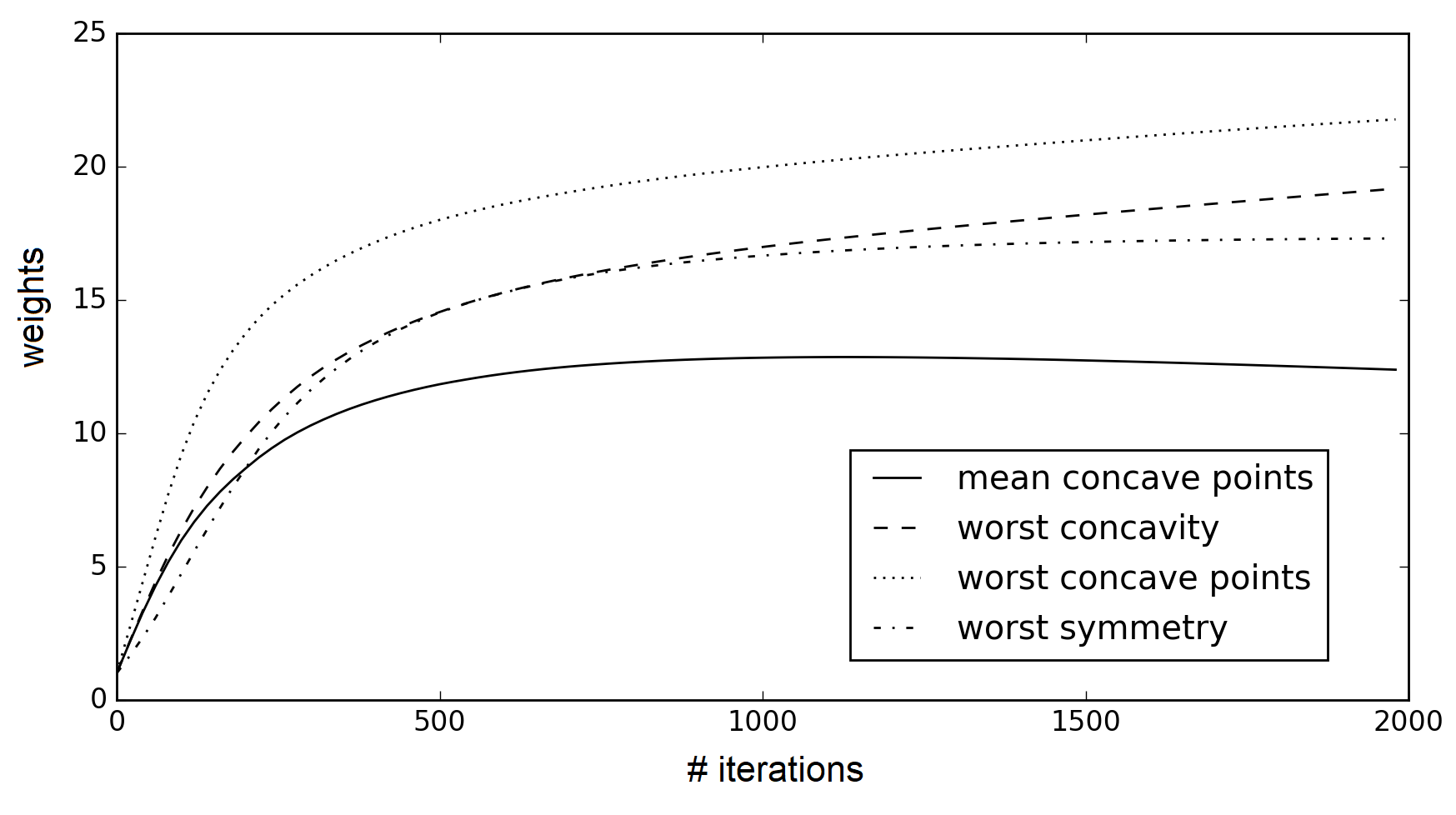}%
\caption{Weights of all features for the Breast Cancer dataset in the case of
the local interpretation using the classification black-box model}%
\label{f:history_bc_clf_loc}%
\end{center}
\end{figure}

\begin{figure}
[ptb]
\begin{center}
\includegraphics[
%%=8.127400in,
%%=10.850600in,
height=2.7404in,
width=3.6529in
]%
{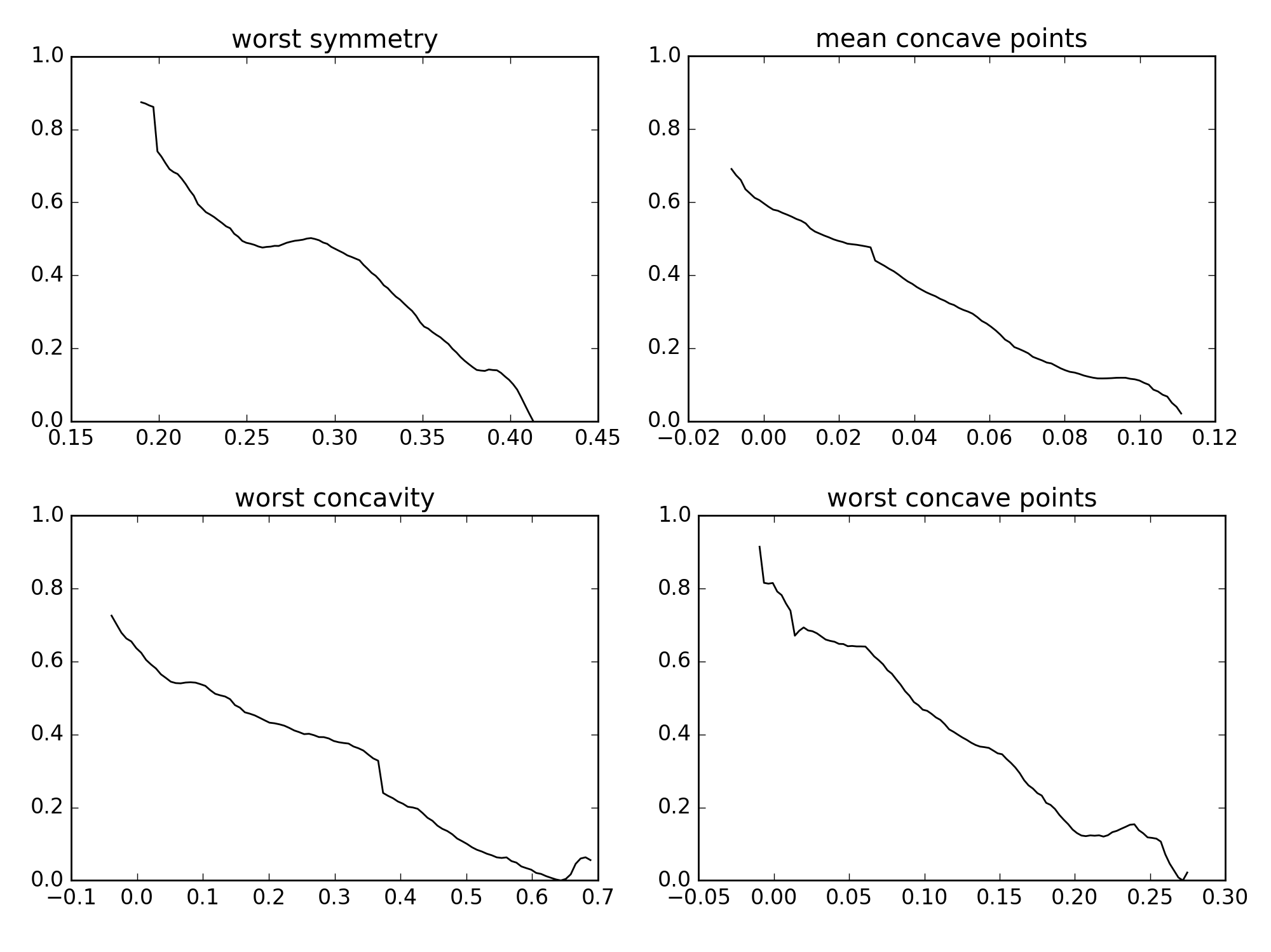}%
\caption{Shape functions of four important features learned on the Breast
Cancer dataset (the x-axis indicates values of each feature, the y-axis
indicates the feature contriburion) for the local interpretation using the
regression black-box model}%
\label{f:featurewise_bc_clf_loc}%
\end{center}
\end{figure}

The importance values of features obtained by using the proposed algorithm are
depicted in Fig. \ref{f:importances_bc_loc}. It can be seen from Fig.
\ref{f:importances_bc_clf_loc} that the same features as in the example with
the regression black-box, including \textquotedblleft worst
symmetry\textquotedblright, \textquotedblleft mean concave
points\textquotedblright, \textquotedblleft worst concavity\textquotedblright,
\textquotedblleft worst concave points\textquotedblright,\ are of the highest
importance. Fig. \ref{f:history_bc_clf_loc} shows how the weights are changed
with increase of the number of iterations $T$ for the above important
features. Individual shape functions for these important features are plotted
in Fig. \ref{f:featurewise_bc_clf_loc}. They are similar to those computed for
the regression black-box model.%

%TCIMACRO{\FRAME{ftbpFU}{4.0811in}{3.4039in}{0pt}{\Qcb{Importance of all
%features for the Breast Cancer dataset in the case of the local interpretation
%using the classification black-box model}}{\Qlb{f:importances_bc_clf_loc}%
%}{importances_bc_clf_loc.png}{\special{ language "Scientific Word";
%type "GRAPHIC";  maintain-aspect-ratio TRUE;  display "USEDEF";
%valid_file "F";  width 4.0811in;  height 3.4039in;  depth 0pt;
%original-width 10.8506in;  original-height 9.0381in;  cropleft "0";
%filename 'importances_bc_clf_loc.png';file-properties "XNPEU";}} }%
%BeginExpansion
%EndExpansion
%

%TCIMACRO{\FRAME{ftbpFU}{3.7727in}{2.1188in}{0pt}{\Qcb{Weights of all features
%for the Breast Cancer dataset in the case of the local interpretation using
%the classification black-box model}}{\Qlb{f:history_bc_clf_loc}}%
%{history_bc_clf_loc.png}{\special{ language "Scientific Word";
%type "GRAPHIC";  maintain-aspect-ratio TRUE;  display "USEDEF";
%valid_file "F";  width 3.7727in;  height 2.1188in;  depth 0pt;
%original-width 9.4777in;  original-height 5.306in;  cropleft "0";
%filename 'history_bc_clf_loc.png';file-properties "XNPEU";}} }%
%BeginExpansion
%EndExpansion
%

%TCIMACRO{\FRAME{ftbpFU}{3.6529in}{2.7404in}{0pt}{\Qcb{Shape functions of four
%important features learned on the Breast Cancer dataset (the x-axis indicates
%values of each feature, the y-axis indicates the feature contriburion) for the
%local interpretation using the regression black-box model}}%
%{\Qlb{f:featurewise_bc_clf_loc}}{featurewise_bc_clf_loc.png}%
%width 3.6529in;  height 2.7404in;  depth 0pt;  original-width 10.8506in;
%original-height 8.1274in;  cropleft "0";  croptop "1";  cropright "1";
%filename 'featurewise_bc_clf_loc.png';file-properties "XNPEU";}} }%
%BeginExpansion
%EndExpansion

\section{Conclusion}

A new interpretation method which is based on applying an ensemble of parallel
GBMs has been proposed for interpreting black-box models. The interpretation
algorithm implementing the method learns a linear combination of GBMs such
that a single feature is processed by each GBM. GBMs use randomized decision
trees of depth 1 as base models. In spite of this simple implementation of the
base models, they show correct results whose intuitive interpretation
coincides with the obtained interpretation. Various numerical experiments with
synthetic and real data have illustrated the advantage of the proposed method.

The method is based on the parallel and independent usage of GBMs during one
iteration. Indeed, each GBM deals with a single feature. However, the
architecture of the proposed algorithm leads to the idea to develop a method
which could consider combinations of features. In this case, the feature
correlation can be taken into account. This is a direction for further
research. Another interesting direction for research is to consider a
combination of the NAMs \cite{Agarwal-etal-20} with the proposed algorithm.

The proposed algorithm consists of several components, including, the Lasso
method, a specific scheme for updating weights. Every component can be
replaced by another implementation which could lead to better results. The
choice of the best configuration for the algorithm is also an important
direction for further research.

The proposed method is efficient mainly for tabular data. However, it can be
also adapted to the image processing which has some inherent peculiarities.
The adaptation is another interesting direction for research in future.

\bibliographystyle{plain}
\bibliography{Boosting,Classif_bib,Deep_Forest,Explain,Explain_med,Lasso,MYBIB,MYUSE,Robots,Survival_analysis}

\end{document}